\documentclass[10pt,twocolumn,letterpaper]{article}
\usepackage{wacv}
\usepackage{times}
\usepackage{epsfig}
\usepackage{graphicx}
\usepackage{amsmath}
\usepackage{amssymb}
\usepackage{booktabs}
\usepackage{tabularx}
\usepackage{subfig}

\def\wacvPaperID{689} 

\wacvalgorithmstrack   

\wacvfinalcopy 

\ifwacvfinal
\usepackage[breaklinks=true,bookmarks=false]{hyperref}
\else
\usepackage[pagebackref=true,breaklinks=true,colorlinks,bookmarks=false]{hyperref}
\fi

\usepackage[capitalize]{cleveref}
\crefname{section}{Sec.}{Secs.}
\Crefname{section}{Section}{Sections}
\Crefname{table}{Table}{Tables}
\crefname{table}{Tab.}{Tabs.}

\usepackage{tabularx}
\usepackage{amsmath}
\usepackage{amssymb}
\usepackage{mathtools}
\usepackage{bm}
\usepackage{overpic}
\usepackage{enumitem} 
\usepackage{overpic} 
\usepackage{color}
\usepackage[accsupp]{axessibility}  
\usepackage[square, comma, sort&compress, numbers]{natbib}
\definecolor{turquoise}{cmyk}{0.65,0,0.1,0.3}
\definecolor{purple}{rgb}{0.65,0,0.65}
\definecolor{dark_green}{rgb}{0, 0.5, 0}
\definecolor{orange}{rgb}{0.8, 0.6, 0.2}
\definecolor{red}{rgb}{0.8, 0.2, 0.2}
\definecolor{darkred}{rgb}{0.6, 0.1, 0.05}
\definecolor{blueish}{rgb}{0.0, 0.3, .6}
\definecolor{light_gray}{rgb}{0.7, 0.7, .7}
\definecolor{pink}{rgb}{1, 0, 1}
\definecolor{greyblue}{rgb}{0.25, 0.25, 1}

\usepackage{blindtext}

\renewcommand{\paragraph}[1]{\vspace{1em}\noindent\textbf{#1}.}

\makeatletter
\let\save@mathaccent\mathaccent
\newcommand*\if@single[3]{%
  \setbox0\hbox{${\mathaccent"0362{#1}}^H$}%
  \setbox2\hbox{${\mathaccent"0362{\kern0pt#1}}^H$}%
  \ifdim\ht0=\ht2 #3\else #2\fi
  }
\newcommand*\rel@kern[1]{\kern#1\dimexpr\macc@kerna}
\newcommand*\widebar[1]{\@ifnextchar^{{\wide@bar{#1}{0}}}{\wide@bar{#1}{1}}}
\newcommand*\wide@bar[2]{\if@single{#1}{\wide@bar@{#1}{#2}{1}}{\wide@bar@{#1}{#2}{2}}}
\newcommand*\wide@bar@[3]{%
  \begingroup
  \def\mathaccent##1##2{%
    \let\mathaccent\save@mathaccent
    \if#32 \let\macc@nucleus\first@char \fi
    \setbox\z@\hbox{$\macc@style{\macc@nucleus}_{}$}%
    \setbox\tw@\hbox{$\macc@style{\macc@nucleus}{}_{}$}%
    \dimen@\wd\tw@
    \advance\dimen@-\wd\z@
    \divide\dimen@ 3
    \@tempdima\wd\tw@
    \advance\@tempdima-\scriptspace
    \divide\@tempdima 10
    \advance\dimen@-\@tempdima
    \ifdim\dimen@>\z@ \dimen@0pt\fi
    \rel@kern{0.6}\kern-\dimen@
    \if#31
      \overline{\rel@kern{-0.6}\kern\dimen@\macc@nucleus\rel@kern{0.4}\kern\dimen@}%
      \advance\dimen@0.4\dimexpr\macc@kerna
      \let\final@kern#2%
      \ifdim\dimen@<\z@ \let\final@kern1\fi
      \if\final@kern1 \kern-\dimen@\fi
    \else
      \overline{\rel@kern{-0.6}\kern\dimen@#1}%
    \fi
  }%
  \macc@depth\@ne
  \let\math@bgroup\@empty \let\math@egroup\macc@set@skewchar
  \mathsurround\z@ \frozen@everymath{\mathgroup\macc@group\relax}%
  \macc@set@skewchar\relax
  \let\mathaccentV\macc@nested@a
  \if#31
    \macc@nested@a\relax111{#1}%
  \else
    \def\gobble@till@marker##1\endmarker{}%
    \futurelet\first@char\gobble@till@marker#1\endmarker
    \ifcat\noexpand\first@char A\else
      \def\first@char{}%
    \fi
    \macc@nested@a\relax111{\first@char}%
  \fi
  \endgroup
}
\makeatother

\newcommand*\basemesh[1]{\widebar{#1}}

\begin{document}

\title{Mesh-Tension Driven Expression-Based Wrinkles for Synthetic Faces}

\author{
Chirag Raman\\
{\small Delft University of Technology}\\
{\tt\small \href{https://chiragraman.com}{chiragraman.com}}
\and
Charlie Hewitt\\
{\small Microsoft}\\
{\tt\small \href{https://chewitt.me}{chewitt.me}}
\and
Erroll Wood\\
{\small Microsoft}\\
{\tt\small \href{https://errollw.com}{errollw.com}}
\and
Tadas Baltru\v{s}aitis\\
{\small Microsoft}\\
{\tt\small tabaltru@microsoft.com}
}
\maketitle
\thispagestyle{empty}
\begin{abstract}
Recent advances in synthesizing realistic faces have shown that synthetic training data can replace real data for various face-related computer vision tasks. A question arises: how important is realism? Is the pursuit of photorealism excessive?  In this work, we show otherwise. We boost the realism of our synthetic faces by introducing dynamic skin wrinkles in response to facial expressions, and observe significant performance improvements in downstream computer vision tasks. Previous approaches for producing such wrinkles either required prohibitive artist effort to scale across identities and expressions, or were not capable of reconstructing high-frequency skin details with sufficient fidelity.
Our key contribution is an approach that produces realistic wrinkles across a large and diverse population of digital humans. 
Concretely, we formalize the concept of mesh-tension and use it to aggregate possible wrinkles from high-quality expression scans into albedo and displacement texture maps.
At synthesis, we use these maps to produce wrinkles even for expressions not represented in the source scans. 
Additionally, to provide a more nuanced indicator of model performance under deformations resulting from compressed expressions, we introduce the 300W-winks evaluation subset and the Pexels dataset of closed eyes and winks. 

\end{abstract}
\section{Introduction}
\label{sec:intro}
\begin{figure}[t]
\centering
\includegraphics[width=\linewidth]{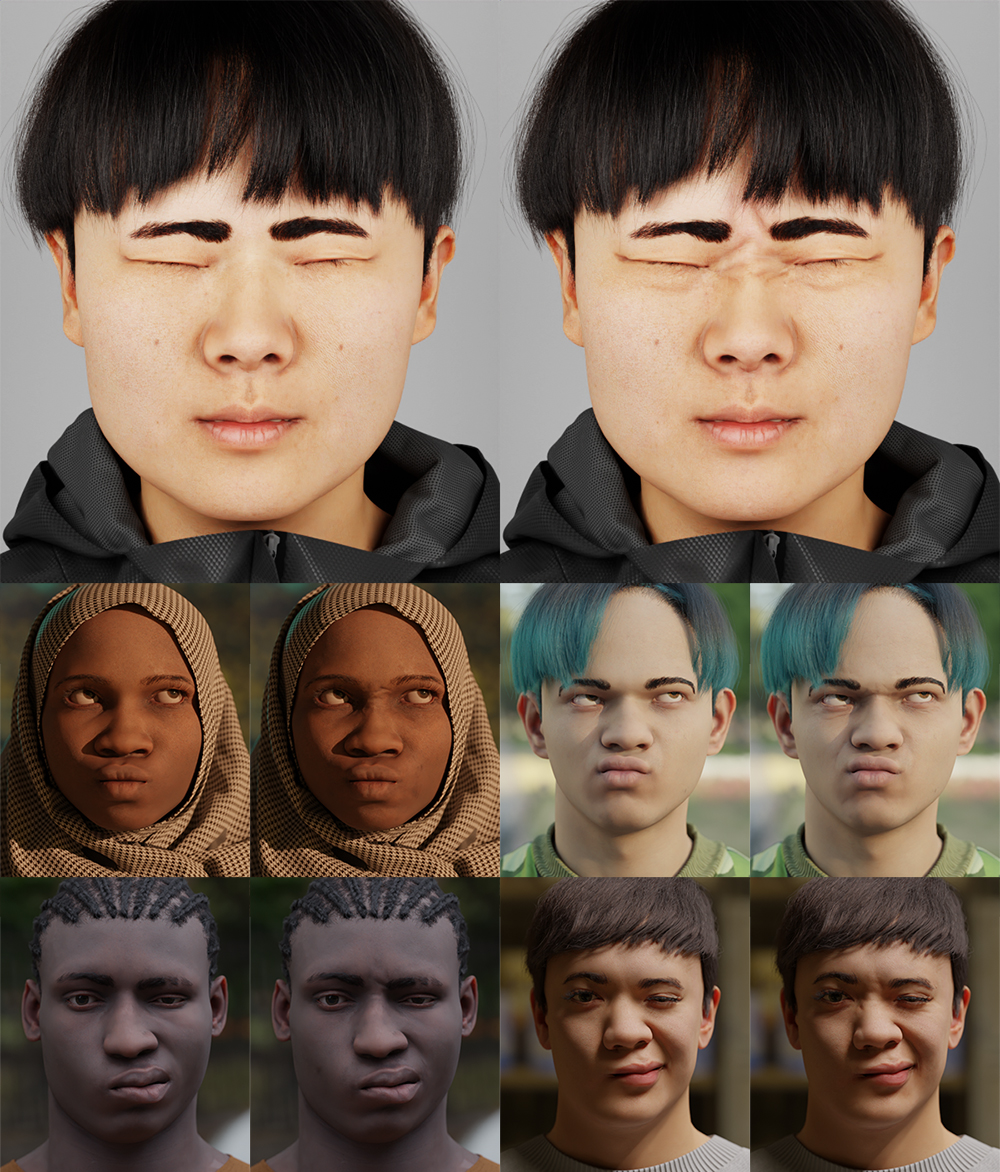}
\caption{
\textbf{Final renders for a diverse set of synthetic identities and expressions.} For each identity we illustrate renders using the base method of \citet{woodFakeItTill2021} (left), and our added technique for generating expression-based wrinkling effects (right). For the same expression parameters, our method produces varied wrinkling effects across distinct identities (middle and bottom row).
}
\label{fig:teaser}
\end{figure}
Synthetic data has been commonly employed for a variety of computer vision tasks including object recognition \cite{yaoSimulatingContentConsistent2020, hodanPhotorealisticImageSynthesis2019, qiuUnrealCVVirtualWorlds2017, rozantsevRenderingSyntheticImages2015}, scene understanding \cite{karMetaSimLearningGenerate2019a, gaidonVirtualWorldsProxy2016, richterPlayingDataGround2016, rosSYNTHIADatasetLarge2016}, eye tracking \cite{woodRenderingEyesEyeShape2015, swirskiRenderingSyntheticGround2014}, hand tracking \cite{muellerGANeratedHandsRealtime2017, simonHandKeypointDetection2017}, and full body analysis \cite{varolLearningSyntheticHumans2017, shottonRealtimeHumanPose2011, huazhongningDiscriminativeLearningVisual2008}. However, the complexity of modeling the human head has largely precluded the generation of full-face synthetics for face-related machine learning. While realistic digital humans have been created for movies and video games, they usually entail significant artist effort per character \cite{hendlerAvengersCapturingThanos2018, karisbrianDigitalHumansCrossing2016}. Consequently in literature, the synthesis of facial training data has been accompanied by simplifications, or a focus on parts of the face such as the eye region \cite{woodLearningAppearancebasedGaze2016, suganoLearningbySynthesisAppearanceBased3D2014} or the \textit{hockey mask} \cite{selaUnrestrictedFacialGeometry2017, richardsonLearningDetailedFace2017, richardson3DFaceReconstruction2016, zengDF2NetDenseFineFinerNetwork2019}. This has resulted in a \textit{domain gap}\textemdash a difference in distributions between real and synthetic facial data that makes generalization challenging. Efforts towards bridging this domain gap have mainly utilized domain adaptation to refine synthesized images \cite{shrivastavaLearningSimulatedUnsupervised2017} or domain-adversarial training where models are encouraged to ignore domain differences \cite{ganinDomainAdversarialTrainingNeural2016}. As such, generating realistic face data has been considered so challenging that it is assumed that synthetic data cannot fully replace real data for in-the-wild tasks \cite{shrivastavaLearningSimulatedUnsupervised2017}. 

To directly address the challenge, \citet{woodFakeItTill2021} attempted to minimize the domain gap at the source, by generating synthetic faces with unprecedented realism. Their method procedurally combines a parametric 3D face model with a comprehensive library of high-quality artist-created assets including textures, hair, and clothing. 
In doing so, the method overcomes a key bottleneck in techniques employed by the Visual Effects (VFX) industry for synthesizing realistic humans\textemdash that of scale. The procedural sampling can randomly create and render novel 3D faces without manual intervention. Machine learning systems trained on the synthesized data for landmark localization and face parsing achieved performance comparable with the state-of-the-art without using a single real image.

However, one limitation of the method proposed by \citet{woodFakeItTill2021} is the lack of dynamic, expression dependent wrinkles. The method generates textures using only the neutral-expression scans, which remain static for all deformations of the underlying face mesh resulting from expression changes. 
In this work we propose a simple yet effective method for incorporating expression-based wrinkles. Our central idea is to capture complex wrinkling effects for an identity from high-resolution scans of their posed expressions. We store all these possible wrinkles into albedo and displacement textures we refer to as \textit{wrinkle maps}. At synthesis, for any arbitrary expression beyond those represented in the source scans, we blend between the neutral and wrinkle textures using a notion of the \textit{tension} in the face mesh to obtain dynamic wrinkling effects. \autoref{fig:teaser} contrasts the results of our method against the current state-of-the-art (SOTA) approach for face synthetics. We also include animated sequences in the Supplementary Material. 

The term \textit{wrinkle maps} was first used by early VFX approaches to refer to artist-defined bump or normal maps for simulating animated wrinkles \cite{oatAnimatedWrinkleMaps, oatRealTimeWrinkles2007, JIMENEZ2011_GPUPRO2B, dutreveRealTimeDynamicWrinkles2009}. However, these approaches suffer from three drawbacks. First, the bump and normal maps only \textit{simulate} underlying geometry changes; the silhouette and shadows which are of relevance for face related tasks such as landmark localization remain unaffected. Second, the methods do not affect the albedo or diffuse textures. Finally, the most crucial drawback is scale. The methods entail manual definition of wrinkle maps and masks for every blendshape for every character. 
In contrast, our automatic mesh-tension driven method naturally scales with the number of identities and expressions, while incorporating real wrinkles for both albedo and displacement textures from scans. Furthermore, we also handle identities without expression scans, transferring plausible wrinkles from the most similar neutral textures.  

To advance the development of synthetics for face-related tasks, we make the following concrete contributions:
\begin{itemize}[leftmargin=*]
\setlength\itemsep{-.3em}
\item A system for dynamic, expression-based wrinkles that scales easily with increasing identities and expressions.
\item A demonstration of empirical qualitative and quantitative improvement over the SOTA synthetics system on face-keypoint localization and surface-normal estimation.
\item Novel evaluation data and metrics for keypoint localization in the eye region where wrinkles are especially relevant for learning tasks.
\end{itemize}

\section{Background: Synthesizing Faces}
We build upon the work of \citet{woodFakeItTill2021} for synthesizing face images for downstream machine learning tasks. Their method involved sampling from a generative 3D blendshape-based face model learned from 3D scans of $511$ individuals with neutral expression. The sampled face is then \textit{dressed up} with samples from a large collection of hair, clothing, and accessory assets. For each synthesized face, the authors employ three textures that remain fixed across all expressions: one albedo map for skin color; one coarse displacement map to encode scan geometry not captured by the sparsity of the vertex-level identity model; and one meso-displacement map to approximate skin-pore level detail built by high-pass filtering the albedo texture. In contrast, we automatically compute an additional sets of albedo and displacement wrinkle textures from expression scans to support dynamic wrinkling effects.
\begin{figure*}[!t]
\centering
\includegraphics[width=\linewidth]{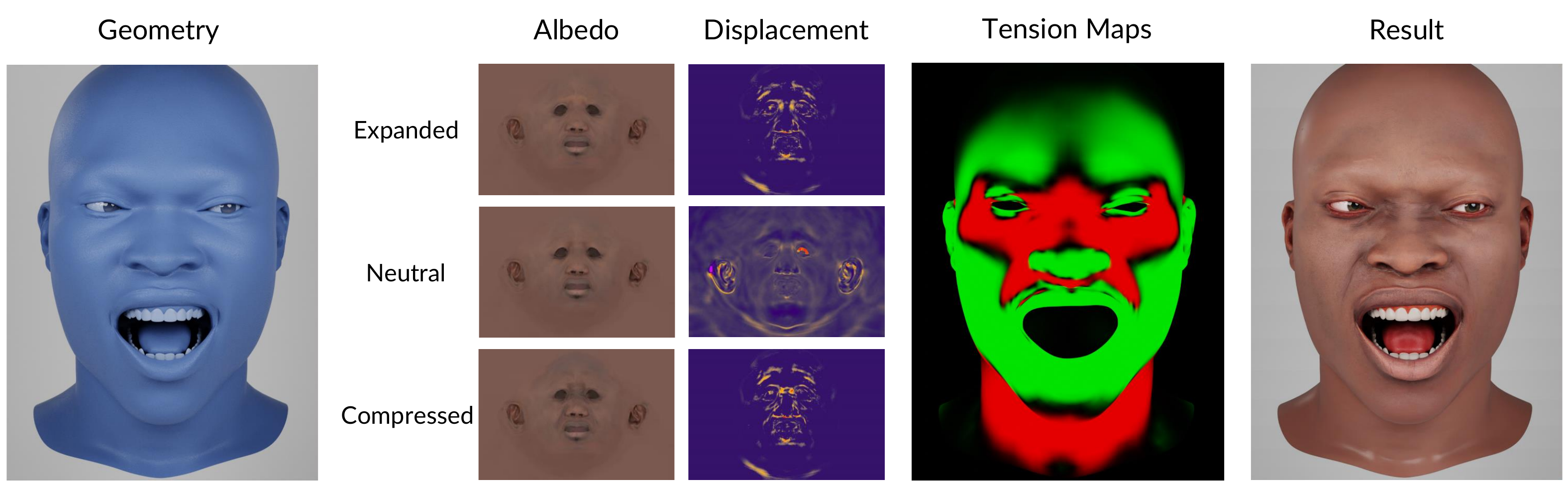}
\caption{
\textbf{Method Overview.}
The state-of-the-art method for face synthetics \cite{woodFakeItTill2021} generates albedo and displacement textures using only the neutral-expression scan for an identity (middle row, also see \autoref{fig:teaser}). In contrast, we automatically compute expanded and compressed texture maps to aggregate wrinkling effects in the face and neck regions across available posed-expression scans for the identity. At synthesis, for a given set of arbitrary expression parameters we compute the local tension at every vertex in the corresponding face mesh: we depict expansion in green and compression in red. This mesh tension serves as weights to dynamically blend between the neutral, expanded, and compressed texture maps to synthesize the wrinkling effect at that vertex. Note that our method can thereby generate wrinkles for expressions even beyond those represented in the source scans.
}
\label{fig:overview}
\end{figure*}

\section{Related Work}
\label{sec:related}

\paragraph{Wrinkle Maps} \citet{oatAnimatedWrinkleMaps} proposed using a pair of bump maps to render animated wrinkles on virtual characters. These bump maps\textemdash called \textit{wrinkle maps}\textemdash store surface normals for an expanded (or stretched) and compressed (or \textit{scrunched-up}) expression, typically obtained from artist sculpted high-resolution meshes. A base normal map stores fine surface details such as pores. In order to achieve independently controlled wrinkles, the face is divided into multiple regions. Each region is specified by an artist-defined mask stored in a texture map. An animated scalar wrinkle weight in the range $[-1, 1]$ then interpolates between the two wrinkle maps for each masked region: at either end of the range one of the wrinkle maps is at its full influence, with a weight of $0$ corresponding to no influence on the base normal map. A similar method was later independently proposed by \citet{duquereisRealtimeSimulationWrinkles2008} using a single wrinkle map. \citet{JIMENEZ2011_GPUPRO2B} expanded on the scheme proposed by \citet{oatAnimatedWrinkleMaps}, allowing for the use of any number of wrinkle maps, with a weight in the range of $[0, 1]$ defining the influence of each map. Subsequent improvements to make the technique amenable in real-time or performance driven settings involved the dynamic generation of either the region masks \cite{dutreveRealTimeDynamicWrinkles2009} or the wrinkle weights \cite{oatRealTimeWrinkles2007}. Both approaches relied on using a \textit{skinned} mesh attached to bones. \citet{dutreveRealTimeDynamicWrinkles2009} proposed generating dynamic region masks by using the bone influence weights from a set of artist defined reference poses. \citet{oatRealTimeWrinkles2007} proposed generating dynamic wrinkle weights by comparing each mesh triangle's area before and after skinning, a technique derived from Microsoft's DirectX 10 Sparse Morph Targets demo \cite{microsoftdirectx10sdkteamSparseMorphTargets2007}. While the term \textit{wrinkle maps} in literature has been alternatively used to refer to bump or normal maps, in this work we use the term to collectively refer to the textures used for synthesizing wrinkles: the albedo and displacement maps corresponding to the expanded and compressed textures.

\paragraph{Simulation Based Approaches} While the use of wrinkle maps is the most common methodology when artistic control is of importance, several alternate techniques have been proposed for simulating wrinkles on 3D surfaces. These methods can broadly be grouped into physical and geometric simulation of wrinkles. An early physical simulation based approach employed a biomechanical perspective, considering the skin as an elastic membrane and modeling the deformations using linear plastic model \cite{wuPhysicallybasedWrinkleSimulation1997}. \citet{boissieuxSimulationSkinAging2000} extended the elastic membrane perspective by modeling the skin as a volumetric substance comprising layers of different materials and using a finite element method for computing deformations. Finite element modeling was also employed in subsequent works to simulate forearm skin wrinkling \cite{Flynn2008FiniteEM}, and skin aging \cite{flynnSimulatingWrinklingAging2010}. \citet{wang2006fast} and \citet{10.1016/j.cag.2005.08.024} proposed energy based approaches. Here,  wrinkle deformations are produced by minimizing an energy function indicating flexure properties of a governing curve on a surface. To produce wrinkles on dynamic meshes such as simulated cloth, \citet{muller2010wrinkle} proposed attaching a higher resolution wrinkle mesh to the coarse base mesh and determining the deviations of the wrinkle mesh vertices using a static solver \cite{muller2007position}. Geometric simulation based approaches typically involve expressing the wrinkles using some geometric primitives. \citet{bandoSimpleMethodModeling2002} represented wrinkles using a cubic Bezier curve, generating their furrows from a sequence of starting points along a user specified direction field. Other proposed techniques involved the use of length preserving constraints on planar curves along with artist placed features at locations on an animated mesh where wrinkling is desired \cite{larbouletteRealTimeDynamicWrinkles2004, liModelingExpressiveWrinkles2007}. \citet{ilieRobustMathematicalModel2012} employed a Hermite spline interpolation along with a modified Rayleigh distribution function to simulate wrinkling activity in facial animations. Subsequent methods extracted wrinkle curves automatically from images \cite{liRealisticWrinkleGeneration2011, vanderfeestenExampleBasedSkinWrinkle2018}. Finally, \citet{guiRealistic3DFacial2016} used both a muscle model and a geometric wrinkle shape function to simulate 3D facial wrinkles. 

\paragraph{Machine Learning Approaches} More recently, several methods for expression and texture synthesis, and facial performance capture have addressed the synthesis of wrinkles. As part of their performance capture system, \citet{caoRealtimeHighfidelityFacial2015} trained regressors for mapping local image appearance to wrinkle displacements to augment a coarse face mesh tracked in real-time. \citet{zengDF2NetDenseFineFinerNetwork2019} and \citet{richardsonLearningDetailedFace2017} proposed convolutional networks based refinement architectures to reconstruct detailed facial geometry from a single image. \citet{naganoPaGANRealtimeAvatars2019} proposed a conditional generative adversarial network architecture for the synthesis of image-based dynamic 3D avatars. Given a single neutral-face input image, their system can generate novel photo-real expressions from alternate viewpoints, including variable details such as wrinkles. More directly, \citet{dengPlausible3DFace2021} proposed a variational autoencoder architecture to synthesize plausible fine-scale wrinkles on a variety of coarse-scale 3D faces.

\section{Synthethizing Expression-Based Wrinkles}
\begin{figure}[t]
\centering
\includegraphics[width=\linewidth]{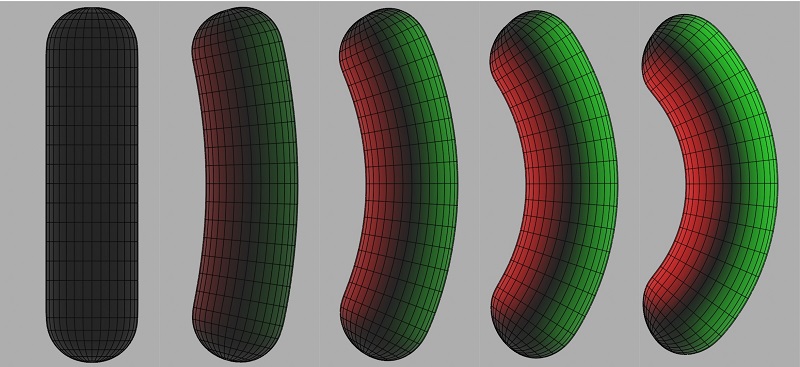}
\caption{
\textbf{Mesh Tension.} We illustrate our computation of mesh tension for various deformations of a simple cylinder. Expansion is depicted in green and compression in red. Black shading corresponds to zero tension.
}
\label{fig:mt-concept}
\end{figure}
\autoref{fig:overview} illustrates an overview of our approach. The underlying idea is that wrinkles can by synthesized additively over the neutral-expression textures. We formalize the concept of mesh tension and use it to automatically aggregate wrinkling effects in a data-driven manner across all expression scans of an identity. 
We store these possible wrinkles corresponding to the expansion and compression deformations of the face in separate albedo and displacement textures, which we collectively refer to as wrinkle maps in this work. Note that displacement maps modify the underlying geometry unlike bump or normal maps that simply simulate the geometry changes. At synthesis, we sample a face mesh from a generative face model \cite{woodFakeItTill2021} and randomly select a set of neutral and wrinkle textures corresponding to an identity from the available scans. We then compute the tension in the face mesh to drive the blending between the neutral and wrinkle maps to obtain dynamic wrinkling effects. 
In contrast with previous learning-based wrinkling methods \cite{richardson3DFaceReconstruction2016, richardsonLearningDetailedFace2017, zengDF2NetDenseFineFinerNetwork2019, dengPlausible3DFace2021}, we do not build a generative model for the textures since such models struggle to reconstruct high frequency details such as wrinkles compared to directly
extracting them from scans. 

\subsection{Mesh Tension}
We formalize mesh tension to 
capture the amount of compression or expansion at each vertex of a 
3D polygon mesh resulting from a deformation. More concretely,
we express mesh tension as a 
function of the mean change in the length of the edges connected 
to a vertex as a result of the deformation. Consider an undeformed mesh $\basemesh{\mathbf{X}} = (\basemesh{V}, \basemesh{E})$ with a sequence of vertices
$\basemesh{V}$ and sequence of edges $\basemesh{E}$, that 
undergoes a deformation to result in the mesh $\mathbf{X} = (V, E)$. We only consider deformations such that
$\basemesh{\mathbf{X}}$ and $\mathbf{X}$ possess the same 
topology. For vertex $v_i \in V$, let  $(e_1,\ldots,e_K)$ denote the 
sequence of $K$ edges  connected to $v_i$, 
with $(\basemesh{e}_1,\ldots,\basemesh{e}_K)$ denoting the 
corresponding edges in $\basemesh{X}$ connected to $\basemesh{v}_i$. We then define the mesh 
tension at $v_i$ as 

\begin{equation}
    t_{v_i} \coloneqq 1 - \frac{1}{K}\sum_{k \in [K]} \frac{\|e_k\|}{\|\basemesh{e}_k\|},
\end{equation}

where $[K] = \{1, \ldots, K\}$, and $\|.\|$ denotes edge length. Note that we subtract from $1$ so that positive values of $t_{v_i}$ indicate compression, negative values indicate expansion, and a value of $0$ indicates no change.
\begin{figure}[t]
\centering
\setkeys{Gin}{width=\linewidth}
\addtolength{\tabcolsep}{-0.5em}
\begin{tabularx}{\linewidth}{@{}*3{>{\centering\arraybackslash}X}@{}}
\includegraphics{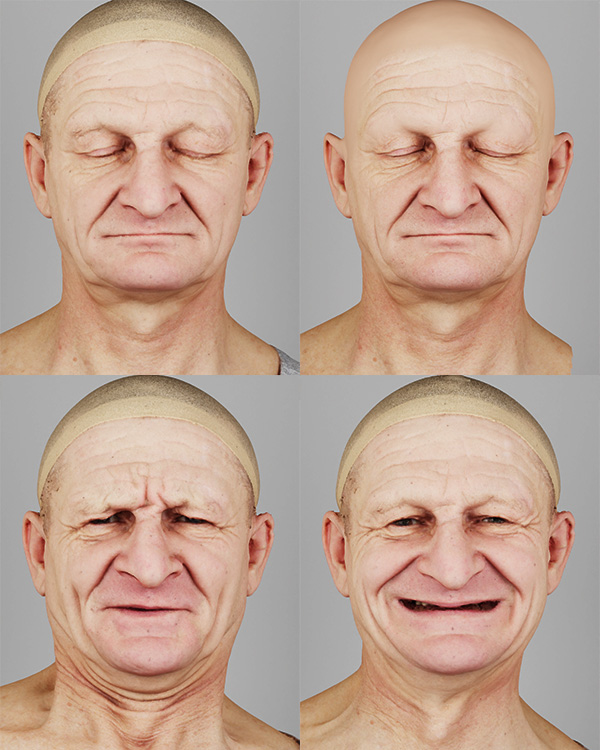} & 
\includegraphics{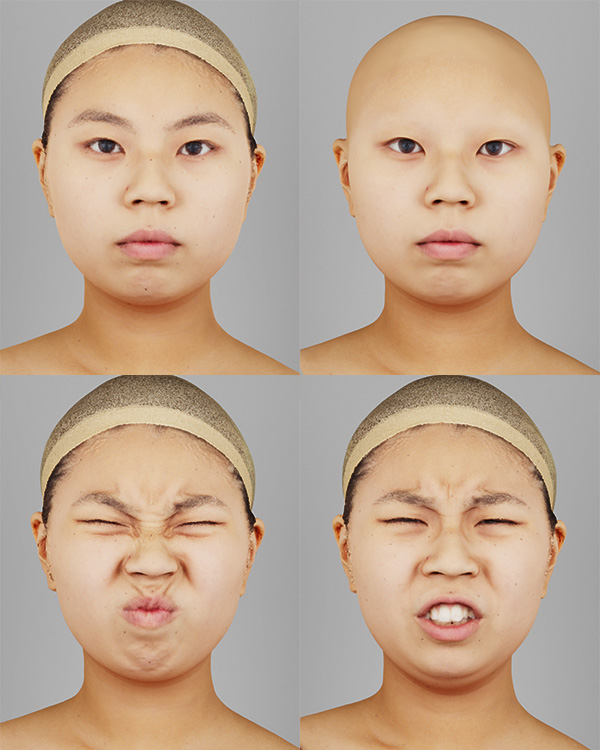} & 
\includegraphics{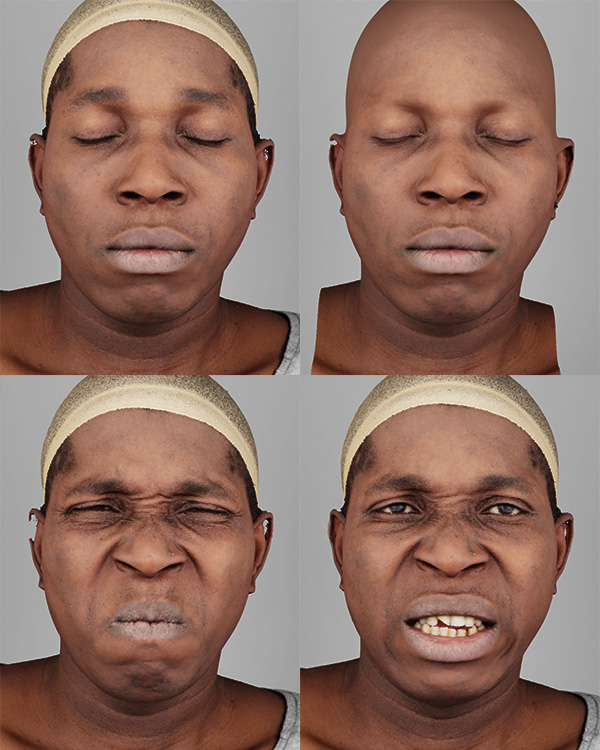}
\end{tabularx}
\caption{
\textbf{Data\textemdash High-resolution 3D Scans.}
For each identity, we illustrate: the raw neutral scan (top-left), the  manually-cleaned neutral scan to remove sensor noise and hair (top-right), and two raw expression scans (bottom).
}
\label{fig:data}
\end{figure}

In practice, for finer manual control we introduce the parameters of strength $s$ to scale the tension, and bias $b$ to artificially favor expansion or compression, computing the weighted tension at $v_i$ as $t'_{v_i} = s\cdot t_{v_i} + b$. Further, we allow for artificial propagation of expansion and compression effects through the mesh. For each effect we introduce a parameter denoting the number of iterations for a morphological dilation  (positive values) or erosion (negative values) operation. The propagation of each effect is first performed independently over the mesh, and the resulting tension values are added for vertices that end up with both expansion and compression. \autoref{fig:mt-concept} illustrates these effects for a simple cylindrical mesh. See \autoref{app:tension} for additional illustrations of the effect of the tension parameters. Code as a Blender \cite{blender} add-on is available at~\url{https://github.com/chiragraman/mesh-tension}

\subsection{Data and Preprocessing}
\begin{figure}[t]
\centering
\setkeys{Gin}{width=\linewidth}
\addtolength{\tabcolsep}{-0.5em}
\begin{tabularx}{\linewidth}{@{}*6{>{\centering\arraybackslash}X}@{}}
\textsf{\scriptsize Raw} & \textsf{\scriptsize Coarse Mask} & \textsf{\scriptsize Fine Mask} & \textsf{\scriptsize Cleaned} & \textsf{\scriptsize Base} \\
\includegraphics{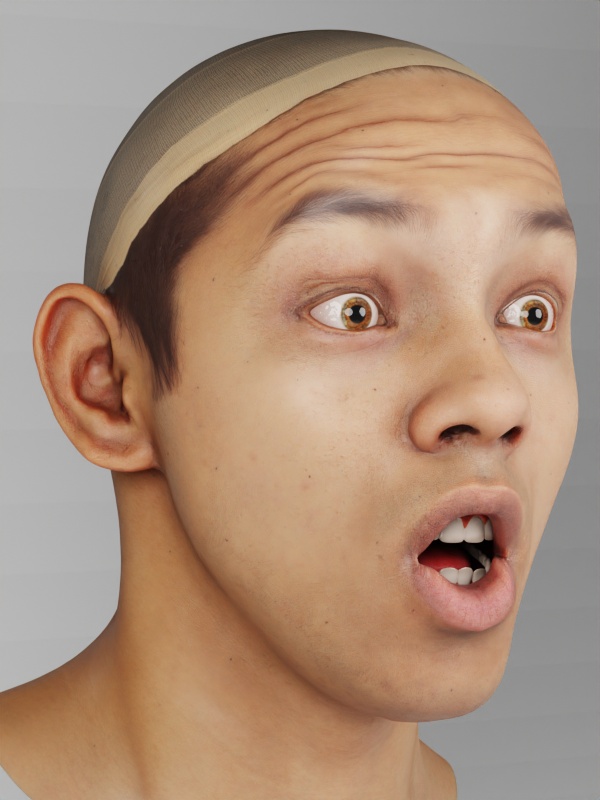} & 
\includegraphics{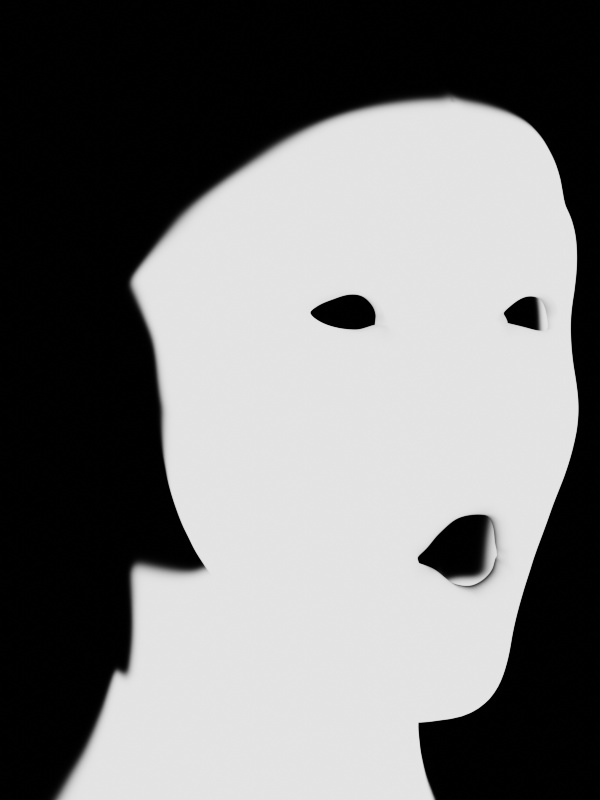} & 
\includegraphics{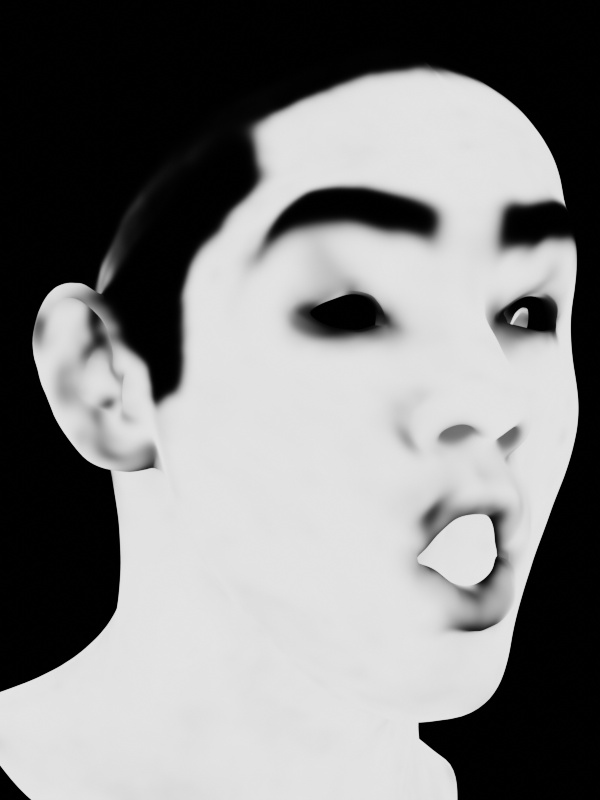} & 
\includegraphics{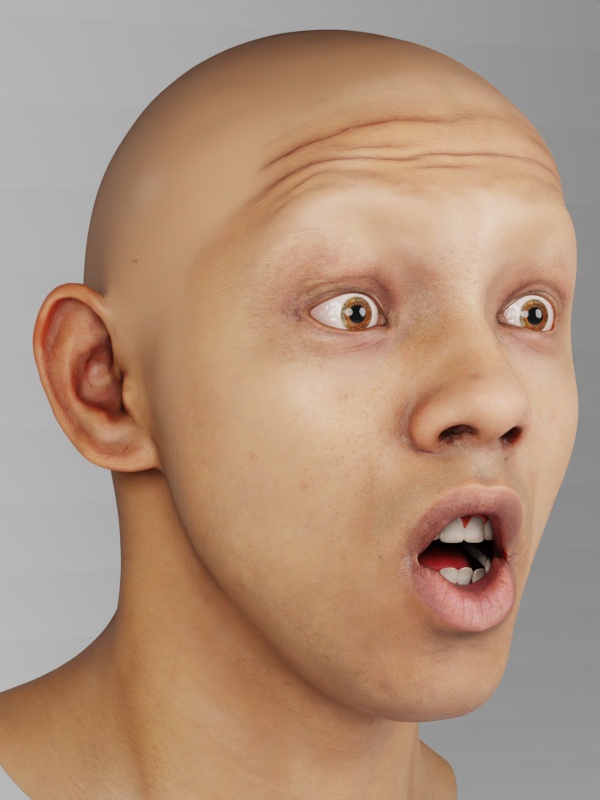} & 
\includegraphics{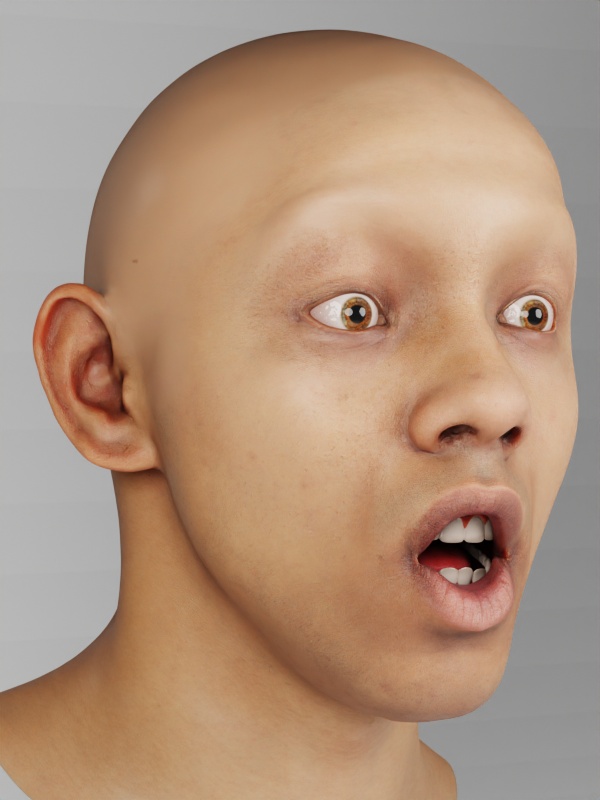} \\
\includegraphics{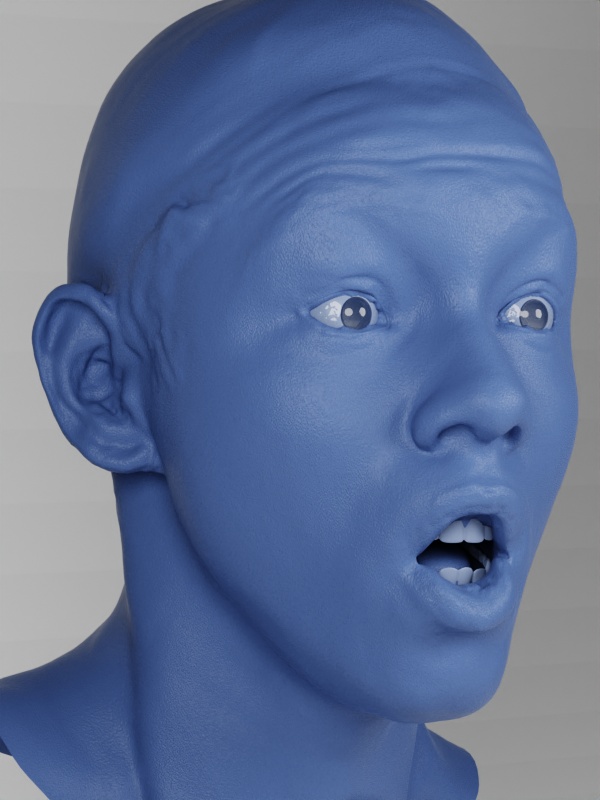} & 
\includegraphics{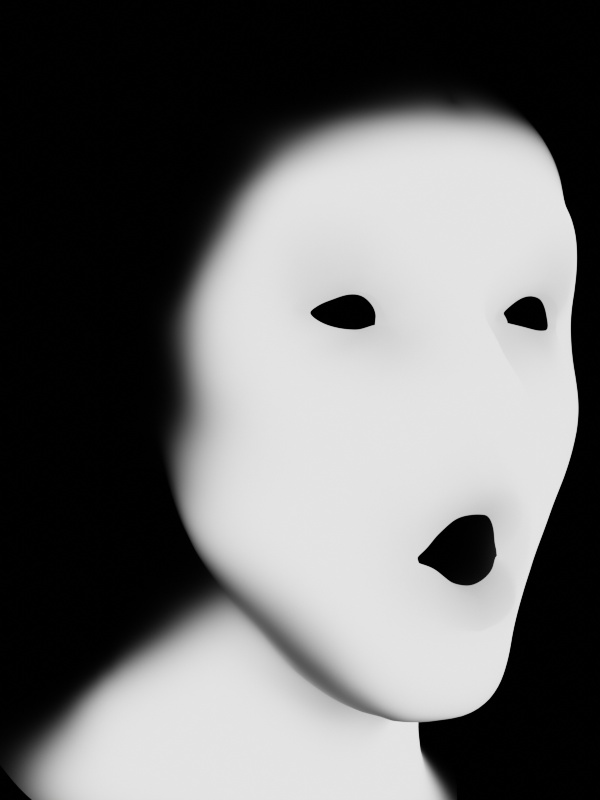} & 
\includegraphics{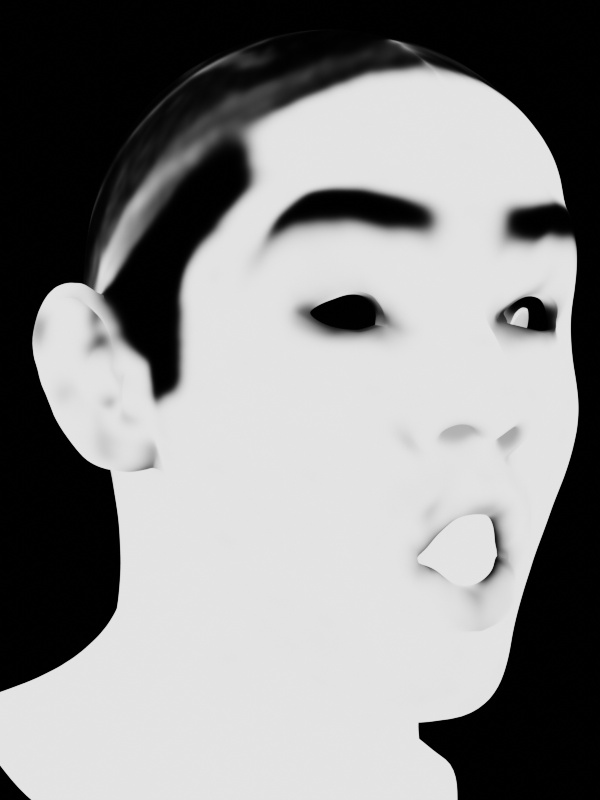} & 
\includegraphics{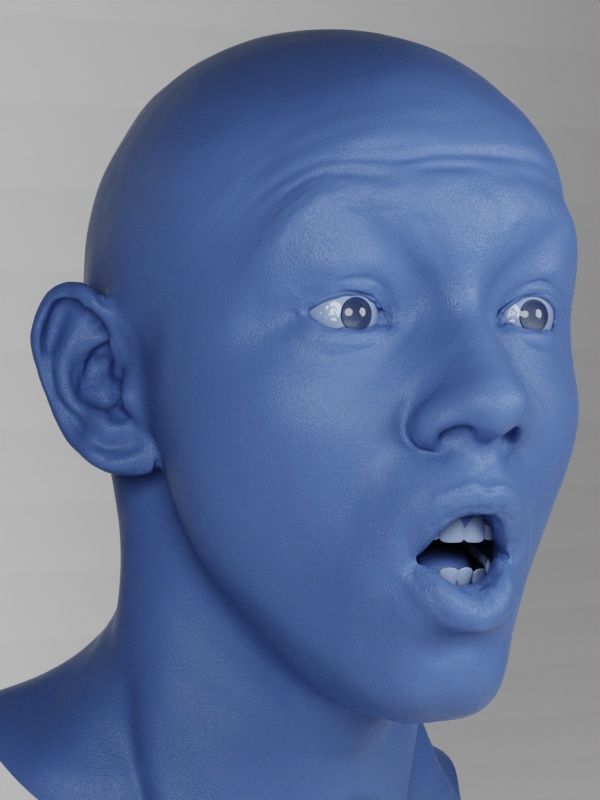} & 
\includegraphics{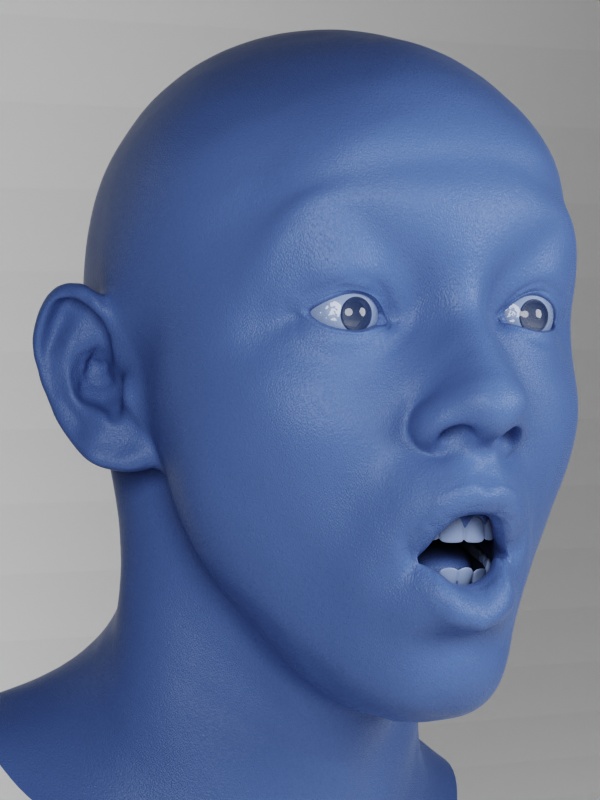} \\
\end{tabularx}
\caption{
\textbf{Cleaning Raw Textures.}
We illustrate the cleanup of albedo (top) and displacement (bottom) textures on the \textit{surprise} expression. We automatically remove the hair and sensor noise artifacts in the raw textures around the head, neck, and cheeks while preserving the desired wrinkles in the nose, forehead, and mouth regions (compared to the base mesh, with neutral albedo and without displacement respectively, for the same expression).
}
\label{fig:cleanup}
\end{figure}
We start with a set of high-quality commercially available 3D scans of $208$ individuals. All $208$ identities contain scans with neutral expressions, while $52$ contain additional scans for posed expressions. The neutral scans were manually cleaned for removing noise and hair artifacts, and registered to the topology of the 3D face model proposed by \citet{woodFakeItTill2021}, resulting in a mesh of $7,667$ vertices and $7,414$ polygons. \autoref{fig:data} illustrates the scans. 

\paragraph{Automatic Cleaning of Expression Scans} The manual cleaning of scans is a labor-intensive process. To automate the process of masking the noise and hair artifacts from the expression scans, we utilize the difference between the raw and manually-cleaned neutral scans. Concretely, we employ a two-stage masking procedure illustrated in \autoref{fig:cleanup}. First, we apply an identity-agnostic coarse mask to filter most artifacts outside of the hockey-mask and neck regions where expression-based wrinkling occurs. Next, to capture the manual changes made by the artists in the cleaning of each neutral scan, we employ a Gaussian Mixture Model-based background subtraction technique \cite{ZIVKOVIC2006773}. Treating the clean neutral textures as background and the raw original ones as foreground, we obtain an identity-specific mask of the noise and hair artifacts for every identity. We apply this fine mask to clean the textures from the corresponding expression scans for each identity.

\subsection{Data-Driven Wrinkle Maps} 
\begin{figure}[t]
\centering
\includegraphics[width=\linewidth]{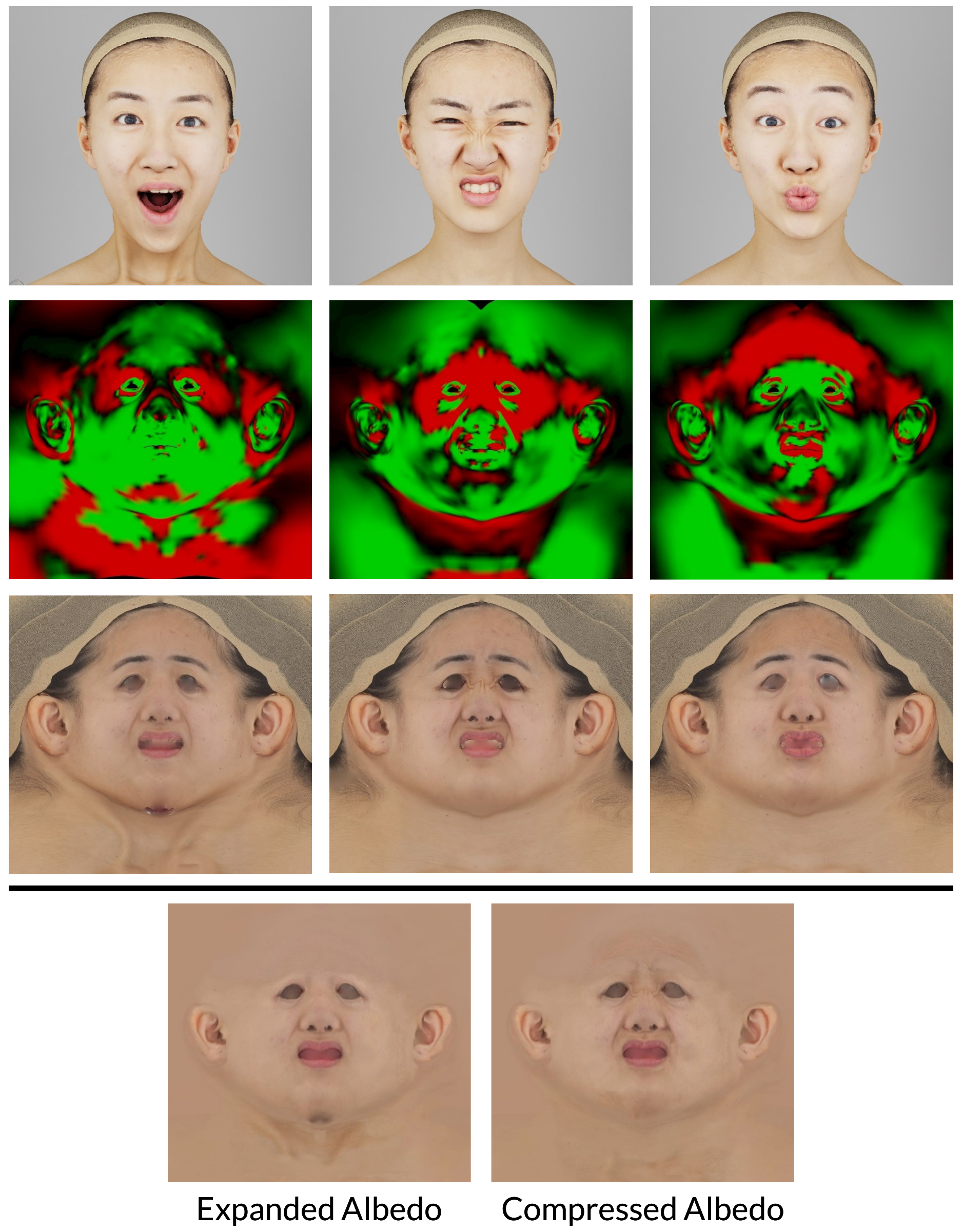}
\caption{
\textbf{Generating Wrinkle Maps from Scans.} We illustrate the computation of albedo wrinkle maps with three raw expression scans (top). We compute the tension maps corresponding to the scans (middle), depicting expansion in green and compression in red. Finally, the expression albedo textures (bottom) are linearly combined using the normalized tension as weights to obtain the expanded and compressed albedo wrinkle maps. A similar procedure is applied to obtain the displacement wrinkle maps. 
}
\label{fig:wrinkle_maps}
\end{figure}
\paragraph{Tension-Weighted Wrinkle Maps} \autoref{fig:wrinkle_maps} illustrates our method for generating wrinkle maps from the face scans. Our underlying idea is to use the tension at each vertex as weights in a linear combination of the cleaned textures across expressions, with zero tension corresponding to the neutral textures. (\autoref{fig:wrinkle_maps} depicts raw textures for easier visual correspondence with the scans.) We begin by fitting the generative face model from \citet{woodFakeItTill2021} to the raw scans and compute the tension maps from the resulting meshes. The individual expansion and compression maps are then normalized using the $\mathrm{softmax}$ function. Finally, we linearly combine expression textures using the normalized tension as weights to obtain the expanded and compressed wrinkle maps. The same procedure is applied to obtain both the albedo and displacement wrinkle maps.

\begin{figure}[t]
\centering
\includegraphics[width=\linewidth]{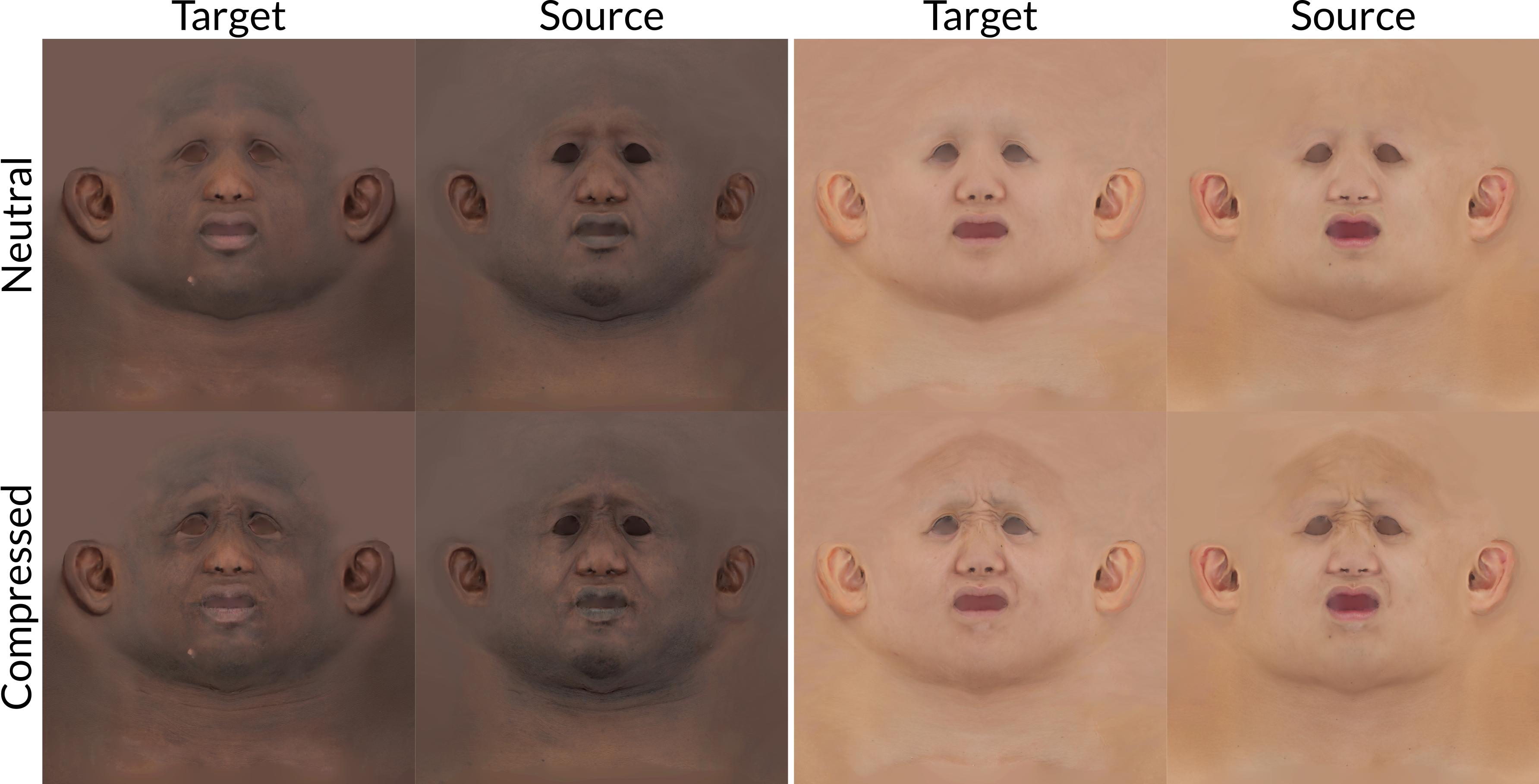}
\caption{
\textbf{Grafting Wrinkles.} For an identity with missing expression scans (target), we find the identity from among those with expression scans that has the most similar neutral albedo map (source). We then graft the wrinkles from the source's wrinkle map onto the target's neutral texture to obtain the target wrinkle maps (here illustrating the compressed albedo). 
}
\label{fig:transfer-method}
\end{figure}

\begin{figure}[t]
\centering
\includegraphics[width=\linewidth]{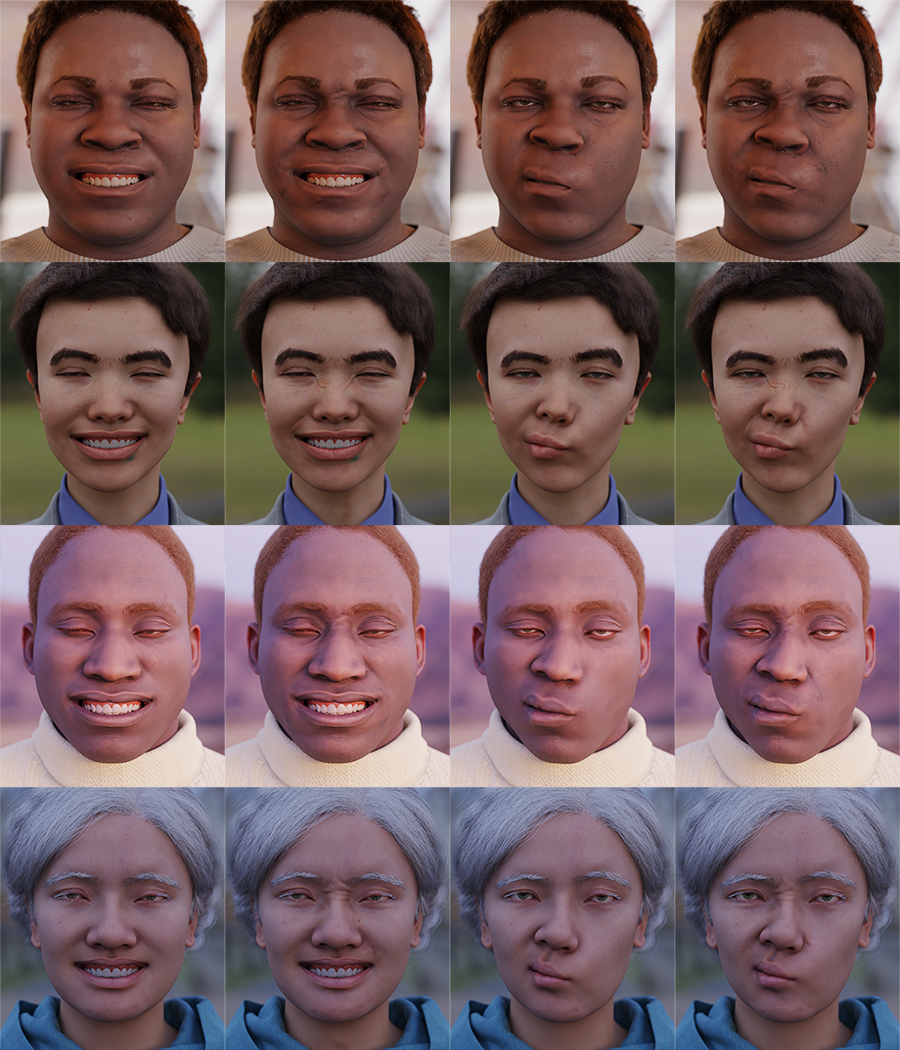}
\caption{
\textbf{Final Renders for Some Identities with Grafted Wrinkles.} We computed the wrinkle maps for these identities by grafting wrinkles from identities with expression scans (see \autoref{fig:transfer-method}). We illustrate two expressions for each identity, without (left) and with wrinkles (right).
}
\label{fig:transfer-results}
\end{figure}

\paragraph{Identities With Missing Expression Scans} How do we compute wrinkle maps for the identities without posed expression scans? We employ a simple wrinkle-grafting procedure. For a target identity without wrinkle maps, we find the source identity with wrinkle maps that has the most similar neutral albedo map, measured by mean squared error in pixel color. For the source identity, we compute the wrinkling effects as the difference between the neutral and wrinkle maps (for both albedo and displacement). We then add this difference to the neutral textures for the target identity to obtain the target wrinkle maps. We illustrate the grafting procedure for the compressed albedo maps in \autoref{fig:transfer-method}, and final example renders with grafted wrinkles in \autoref{fig:transfer-results}.
\section{Experiments and Results}
\begin{table}
\centering
\caption{
\textbf{Landmark Localization on 300W.} We normalize mean error using interocular distance. Lower is better.
} 
\label{tab:lmk-full}
\footnotesize
\begin{tabular*}{\linewidth}{@{}l@{\extracolsep{\fill}}ccc@{}}
\toprule
\textbf{Method}      & \textbf{Common} & \textbf{Challenging} & \textbf{Private} \\
 & \textbf{NME} & \textbf{NME} & \textbf{FR}$_{\mathbf{10\%}}$ \\
\midrule
\textbf{Trained on Real Data}\\
LAB \cite{wu2018look} & $2.98$ & $5.19$ & $0.83$ \\
AWING \cite{wang2019adaptive} & $\mathbf{2.72}$ & $\mathbf{4.52}$ & $\underline{0.33}$ \\
ODN \cite{zhu2019robust} & $3.56$ & $6.67$ & -\\
3FabRec \cite{browatzki2020} & $3.36$ & $5.74$ &  $\mathbf{0.17}$ \\
LUVLi \cite{kumar2020luvli} & $\underline{2.76}$ & $5.16$ & -\\
\midrule
\textbf{Trained on Synthetic Data}\\
No wrinkles \cite{woodFakeItTill2021} & $3.11$ & $4.84$ & $\underline{0.33}$ \\
Ours (wrinkles) & $3.10$  & $\underline{4.83}$ & $\mathbf{0.17}$ \\
\bottomrule
\end{tabular*}
\end{table}

We evaluate our proposed mesh-tension driven wrinkles both quantitatively and qualitatively on two face analysis tasks: landmark detection (Section~\ref{sec:landmark_det}) and normal estimation (Section~\ref{sec:normals}).
We compare adding mesh-tension to the existing SOTA method for full-face synthetics, and compare the performance of models trained on the resulting data against SOTA approaches in the field for these tasks.

\subsection{Landmark Localization}
\label{sec:landmark_det}

\textbf{Experimental Details.} We use direct regression based facial landmark detection~\cite{woodDense2022} with an off-the-shelf ResNet~101~\cite{he2016deep}.
We use a $256\times256$~px RGB image as input to predict $703$ dense facial landmarks.
We additionally employ label translation~\cite{woodFakeItTill2021} to deal with systematic inconsistencies between our $703$ predicted dense landmarks and the $68$ sparse landmarks labeled as ground truth in our evaluation datasets (this is done only for \autoref{tab:lmk-full}).

As a training dataset we rendered $100k$ synthetic images, consisting of $20k$ identities with $5$ frames for each identity (different view-points, expressions, and environments). We also generated ground-truth annotations of $703$ dense 2D landmarks from the face-meshes to accompany each image.
We train our models for $300$ epochs using PyTorch Lightning, starting with a learning rate of $1\mathrm{e}{-3}$ and halved every 100 epochs.

\textbf{Evaluation Datasets and Metrics.} We use the \textbf{300W} dataset \cite{sagonas2016threew} (with common, challenging and private subsets), and employ the standard normalized mean error (NME) and failure rate (FR$_{10\%}$) error metrics~\cite{sagonas2016threew}.

While the 300W dataset provides evaluation of overall landmark detection performance, it is not sensitive enough to detect improvements in specific parts of the face or during particular expressions.
We identify a small subset of $30$ images from 300W that contain winks and compressed face expressions (\textbf{300W-winks}) to provide a more nuanced indication of performance under such deformations.
We report errors for eyelid-landmarks by taking a point-to-line distance from every predicted eyelid landmark to the corresponding polyline defining an eyelid in ground truth. This metric allows us to better understand eye region error and to use different landmark definitions in training and evaluating models (e.g. from our 703 landmark model or from 98 landmark models \cite{wang2019adaptive}). See \autoref{app:300w-winks} for the list of images in 300W-winks.
\begin{table}
\centering
\caption{
\textbf{Landmark Localization - Eyes.} We report eye-opening errors for Pexels, and eyelid point-to-polyline errors for 300W and the \textit{winks} subset. In all cases normalized by bounding-box diagonal. Lower is better.
} 
\label{tab:lmk-eye}
\footnotesize
\begin{tabular*}{\linewidth}{@{}l@{\extracolsep{\fill}}ccc@{}}
\toprule
\textbf{Method} & \textbf{Pexels} & \textbf{300W} & \textbf{300W-winks}\\
\midrule
\textbf{Trained on Real Data}\\
AWING \cite{wang2019adaptive} & 1.06 & 0.62 & $\mathbf{0.69}$ \\
3FabRec \cite{browatzki2020} & 3.60 & 0.81 & 1.32 \\
\midrule
\textbf{Trained on Synthetic Data}\\
No wrinkles \cite{woodFakeItTill2021} & $0.97$ & $0.51$ & $0.86$ \\
Ours (wrinkles) & $\mathbf{0.86}$ & $\mathbf{0.48}$ &  $0.74$ \\
\bottomrule
\end{tabular*}
\end{table}

We also introduce a \textbf{Pexels} dataset which contains 318 images of fully closed eyes (because of blinking, scrunching or compressing the face) and 105 images with only a single eye closed (winking).
This allows us to asses model performance under such conditions which are rare in other datasets.
To collect the data we used a stock photography website \footnote{\url{https://www.pexels.com/}} using search terms \textit{wink/blink/compress/scrunched} and similar image searches. 
We select only semi-frontal images with no or limited occlusion of the eyes to best evaluate performance in that region. 
The URLs of the images selected can be found in \autoref{app:pexels}.
Knowing which images contain fully closed eyes or just a single eye closed allows us to measure eyelid accuracy without explicit landmark annotations.
We define the eye opening error as the mean eye aperture of both eyes in the \textit{eye-closed} case and eye aperture of closed eye in the \textit{wink} case. See \autoref{app:metrics} for illustrations of the above two metrics.

\textbf{Baselines.} We compare against recent SOTA methods trained on images of real faces. For subsequent nuanced analysis on 300W-winks and Pexels we consider the methods of \citet{wang2019adaptive} and \citet{browatzki2020} since they collectively yield the best performance on 300W. 

\begin{figure}[t]
\centering
\setkeys{Gin}{width=\linewidth}
\addtolength{\tabcolsep}{-0.5em}
\begin{tabularx}{\linewidth}{@{}*4{>{\centering\arraybackslash}X}@{}}
\multicolumn{2}{c}{\textsf{\footnotesize Trained on Real Data}} &
\multicolumn{2}{c}{\textsf{\footnotesize Trained on Synthetic Data}} \\
\textsf{\scriptsize AWING \cite{wang2019adaptive}} & 
\textsf{\scriptsize 3FabRec \cite{browatzki2020}} &
\textsf{\scriptsize No Wrinkles \cite{woodFakeItTill2021}} & 
\textsf{\scriptsize Ours (Wrinkles)}\\
\end{tabularx}
\includegraphics[width=\linewidth]{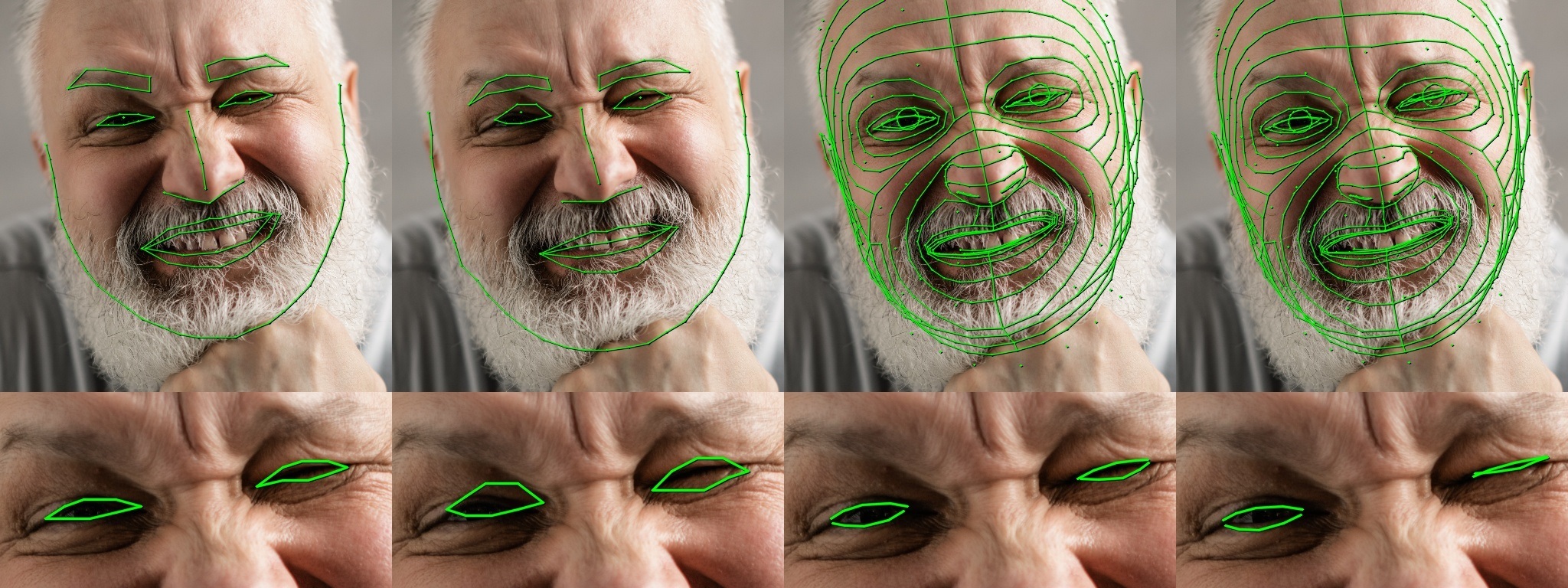}\\
\includegraphics[width=\linewidth]{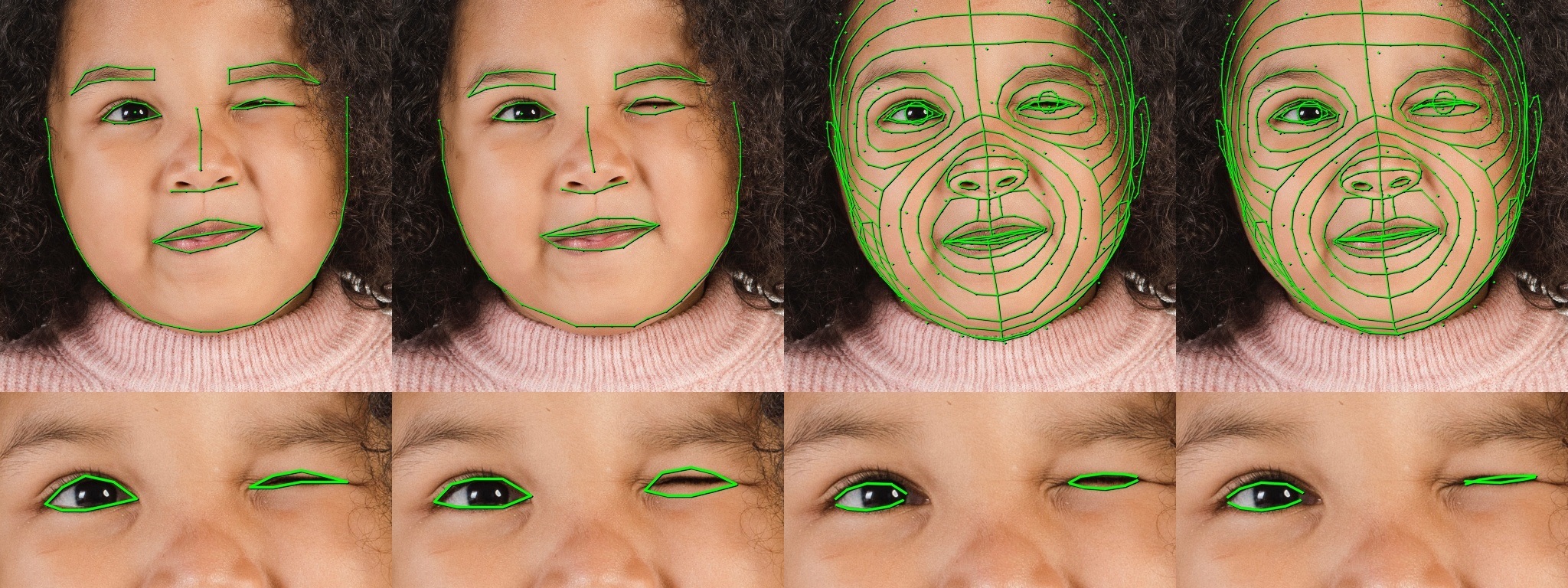}\\
\includegraphics[width=\linewidth]{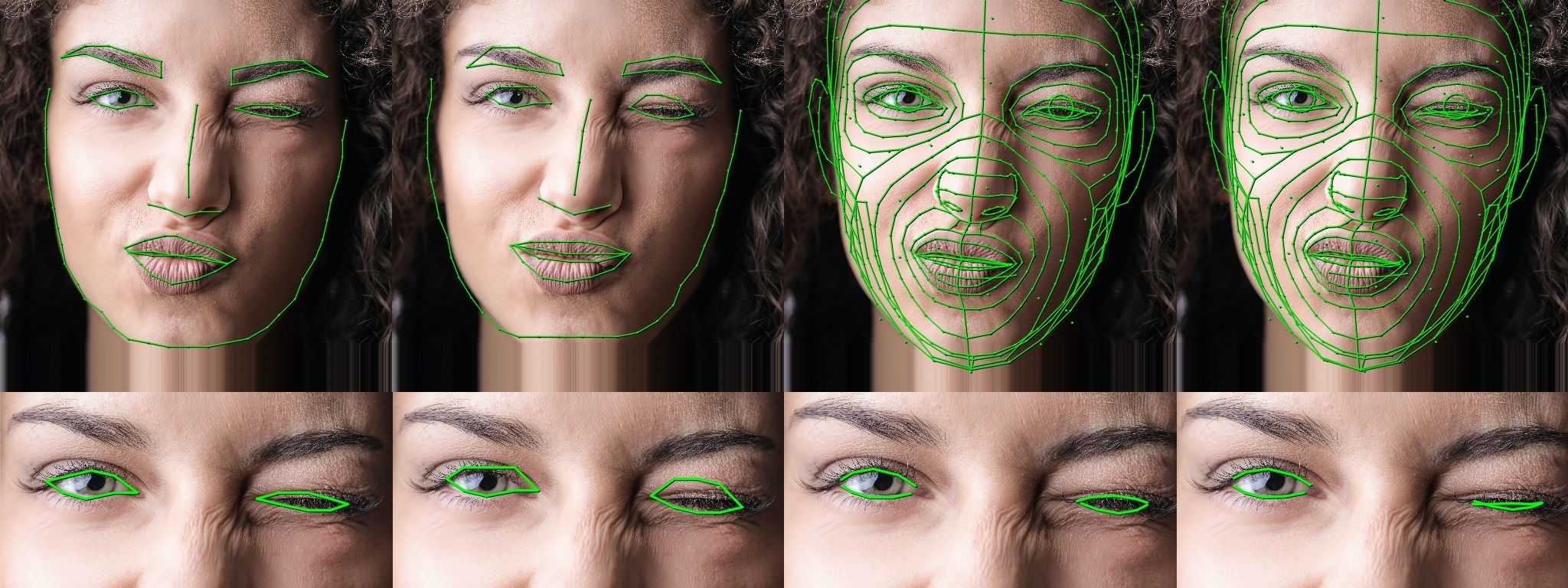}
\caption{
\textbf{Qualitative results for landmark localization on Pexels.} Training on synthetic faces with our expression-based wrinkles is crucial for localizing keypoints in compressed regions of the face.
}
\label{fig:ldmks-eyes}
\end{figure}

\textbf{Results.} From \autoref{tab:lmk-full} we see that our proposed mesh-tension driven wrinkles provide a marginal improvement for landmark localization. However, when we look at specific eye region results on 300W, 300W-winks and Pexels in \autoref{tab:lmk-eye}, we see that improvement is much larger for the eye region and our synthetic-only trained approaches outperform real-data based models. Also see \autoref{fig:ldmks-eyes} and \autoref{app:300W-preds}.

\begin{figure}[t]
\centering
\setkeys{Gin}{width=\linewidth}
\addtolength{\tabcolsep}{-0.5em}
\begin{tabularx}{\linewidth}{@{}*4{>{\centering\arraybackslash}X}@{}}
\textsf{\footnotesize Base} & 
\textsf{\footnotesize Disp. Only} &
\textsf{\footnotesize Albedo Only} & 
\textsf{\footnotesize Albedo \& Disp.}\\
\includegraphics{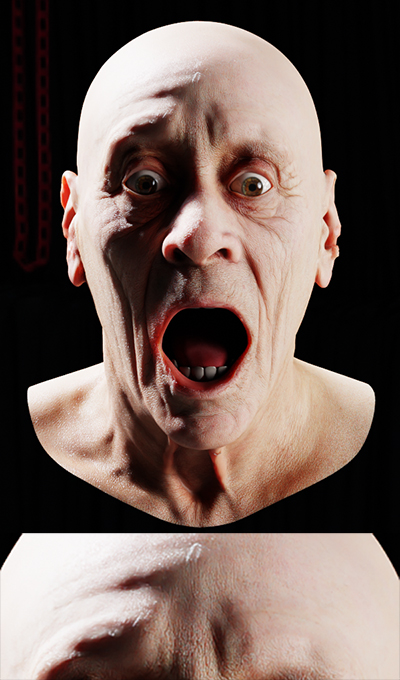} &
\includegraphics{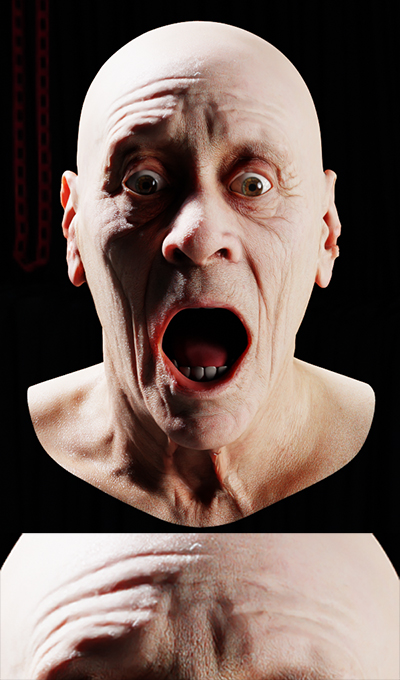} & 
\includegraphics{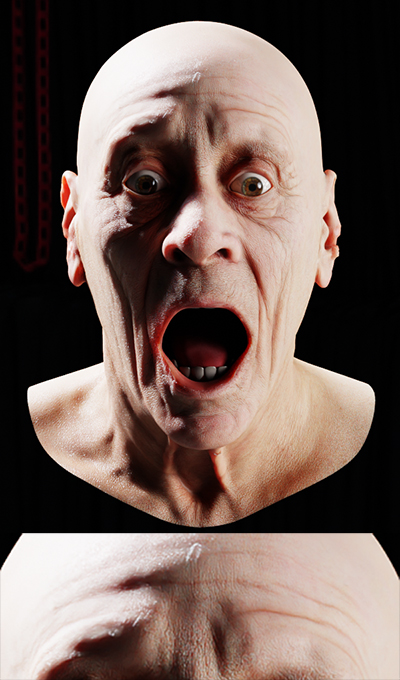} &
\includegraphics{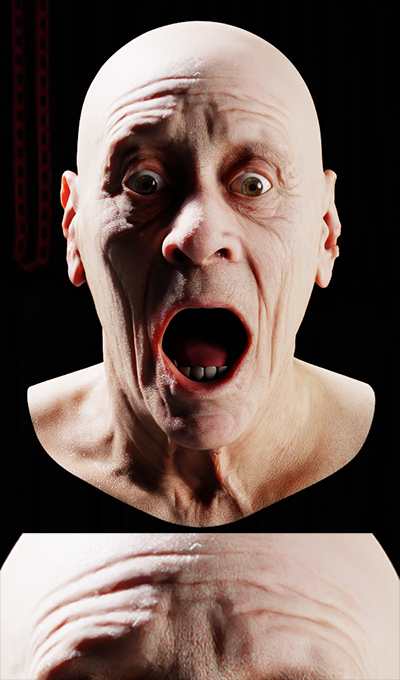}
\end{tabularx}
\caption{
\textbf{Expression-Based Wrinkle Components.} We add wrinkles through two components: displacement and albedo. Here we show each in isolation. Displacement is critical for achieving realistic lighting of wrinkles. Especially note the forehead (zoomed) and neck regions.}
\label{fig:components}
\end{figure}

\textbf{Ablation.} We further analyze the importance of the albedo and displacement wrinkling components for landmark detection.
From \autoref{fig:components} and \autoref{tab:ablation} we see that displacement plays a more important role than albedo in improving performance, but best results are achieved through a combination of both.

\begin{table}
\centering
\caption{
\textbf{Landmark Localization Ablation.} We report eye-opening errors for Pexels, and eyelid point-to-polyline errors for 300W and the \textit{winks} subset. Lower is better.
} 
\label{tab:ablation}
\footnotesize
\begin{tabular*}{\linewidth}{@{}l@{\extracolsep{\fill}}cccc@{}}
\toprule
Dataset & Base & Disp. Only & Albedo Only & Full \\
\midrule
300W & $0.51$ & $0.51$ & $0.50$ & $\mathbf{0.48}$ \\
300W-winks & $0.86$ & $0.76$ & $0.80$ & $\mathbf{0.74}$ \\
Pexels & $0.97$ & $\mathbf{0.86}$ & $0.89$ & $\mathbf{0.86}$ \\
\bottomrule
\end{tabular*}
\vspace{-10pt}
\end{table}

\subsection{Surface-Normals Prediction}
\label{sec:normals}

Surface normals can be used to infer 3D information about a surface from 2D images, and have been used in several human-centered vision tasks such as clothing~\cite{alldieck2019tex2shape} and face-shape~\cite{abrevaya2020cross} reconstruction and relighting~\cite{sengupta2018sfsnet}.

We train a U-Net~\cite{ronneberger2015u} with a ResNet~18~\cite{he2016deep} encoder to predict camera-space surface normals of the face.
As input we use $256\times256$~px RGB images from a dataset of $50k$ synthetics images. 
The network is trained for $200$ epochs using cosine similarity loss with a learning rate of $1\mathrm{e}{-3}$.
Camera-space surface normal images rendered as part of our synthetic data pipeline are used as ground-truth.

\begin{figure}[t]
\centering
\begin{tabularx}{\linewidth}{@{}*4{>{\centering\arraybackslash}X}@{}}
\textsf{\footnotesize Input Image} & \textsf{\footnotesize No Wrinkles \cite{woodFakeItTill2021}} & \textsf{\footnotesize Ours (Wrinkles)} & \textsf{\footnotesize \cite{woodFakeItTill2021} Top Inset / Ours Bottom}\\
\end{tabularx}
\includegraphics[width=\linewidth]{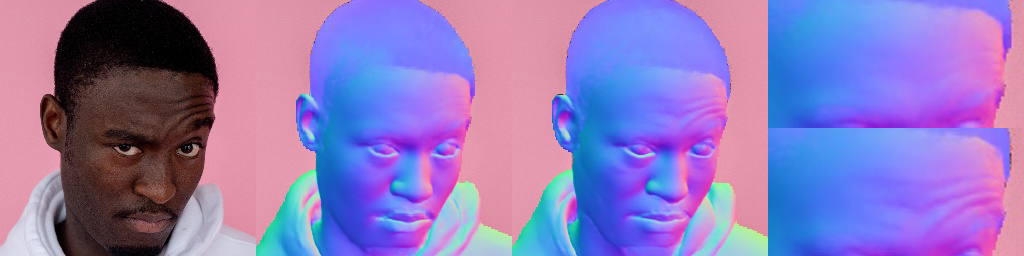}\\
\includegraphics[width=\linewidth]{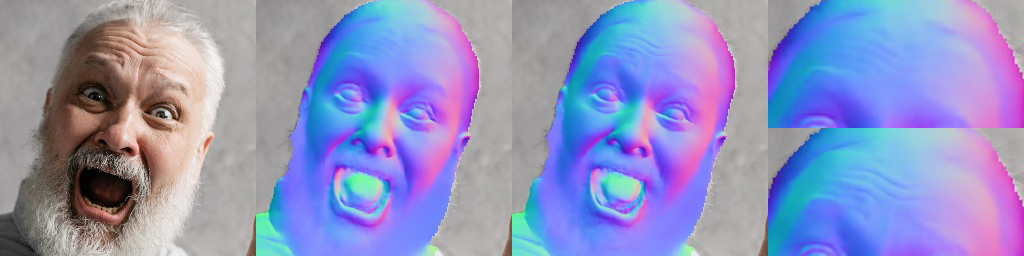}\\
\includegraphics[width=\linewidth]{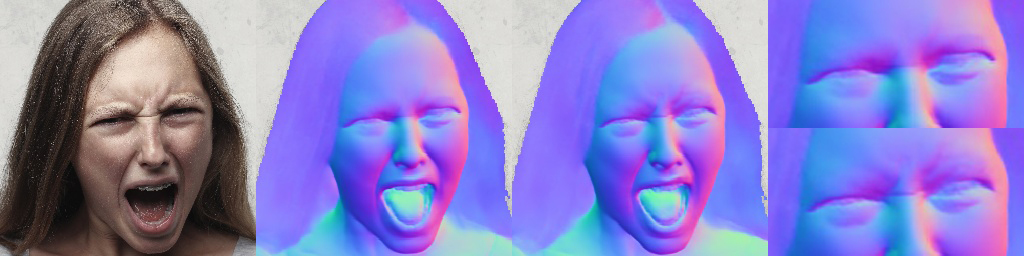}
\caption{
\textbf{Qualitative Surface-Normals Predictions on Pexels.} The model trained on synthetic faces with wrinkles recovers significantly more high-frequency details.}
\label{fig:normals}
\vspace{-10pt}
\end{figure}

\begin{figure}
\renewcommand\tabularxcolumn[1]{m{#1}}
\centering
\begin{tabularx}{\linewidth}{@{}*4{>{\centering\arraybackslash}X}@{}}
\textsf{\footnotesize Input Image} & 
\textsf{\footnotesize CMDFN~\cite{abrevaya2020cross}} & 
\textsf{\footnotesize SfSnet~\cite{sengupta2018sfsnet}} & 
\textsf{\footnotesize Ours}\\
\end{tabularx}
\includegraphics[width=\linewidth]{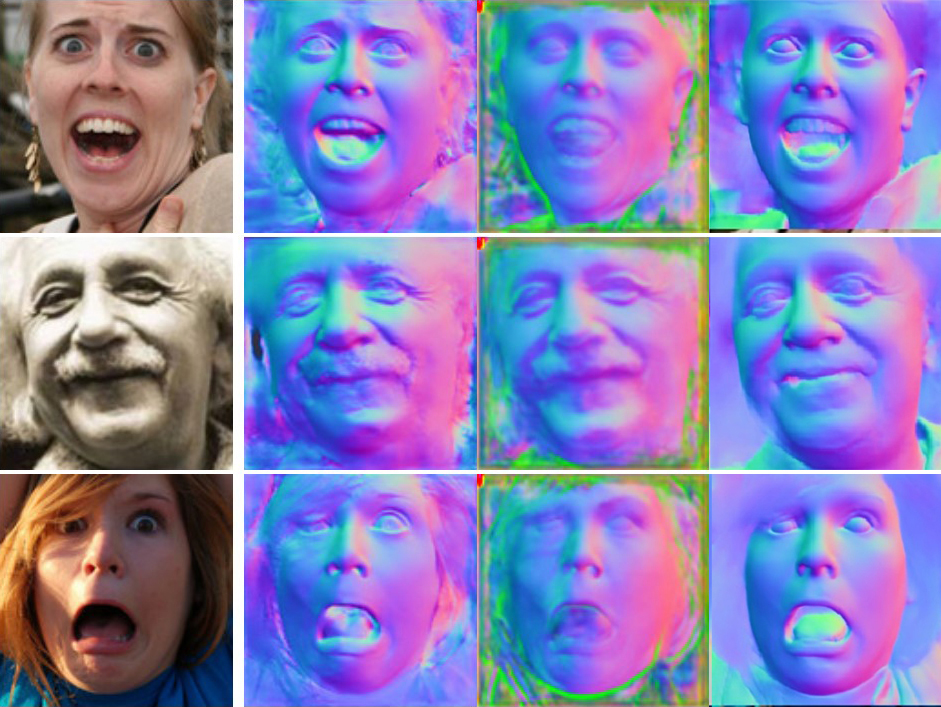}
\caption{
\textbf{Qualitative Comparison against SOTA}. Our synthetic data-only U-Net yields predictions comparable to SOTA while being less noisy and more robust to lighting.}
\label{fig:normals-sota}
\vspace{-10pt}
\end{figure}

Results on real images are shown in \autoref{fig:normals}; the network trained on images synthesized with our method recovers more high-frequency detail on the face. 
As shown in \autoref{fig:normals-sota}, we achieve comparable results to other recent methods for face surface-normals prediction~\cite{abrevaya2020cross,sengupta2018sfsnet}.
Further comparisons are provided in \autoref{app:normals}.
\section{Conclusion}

We have presented a method for introducing dynamic expression-based wrinkles to synthetic faces that yields improved performance on the downstream tasks of landmark localization and surface-normals estimation, especially for regions of the face most deformed by expressions. 

Our use of tension in the face mesh is key in the automatic scaling of our method with identities and expressions, which has been a bottleneck for past wrinkling approaches that rely on prohibitive artist effort. In addition, our data-driven approach also enables the capturing of real wrinkles from scans which doesn't require artistic judgment.  


By boosting the realism of synthesized faces with dynamic wrinkles, we have made an explicit case for synthetic data: our method yields improved performance for models on downstream tasks. 
In addition, synthesizing data with diverse faces across races and genders involves significantly less effort than collecting representative datasets in the wild. Consequently, downstream real-life systems developed using such synthetic data are less likely to suffer from unfair biases along these sensitive variables.  

\section*{Acknowledgments}
Chirag would like to thank: Tom Cashman, Stephan Garbin, and Panagiotis Giannakopoulos for the insightful discussions; Sebastian Dziadzio for help with fitting the face model; Sarah Roberts for being an infallible remover of obstacles; and Steve Miller (BA: @shteeve) for an initial implementation of the Blender mesh tension add on.

{
    \small
    \bibliography{macros,main}
}

\appendix
\onecolumn
\begin{center}
\Large
\textbf{Mesh-Tension Driven Expression-Based Wrinkles for Synthetic Faces} \\
\vspace{0.5em}Appendices \\
\vspace{1.0em}
\normalsize
\author{
Chirag Raman$^1$\qquad Charlie Hewitt$^2$\qquad Erroll Wood$^2$\qquad Tadas Baltru\v{s}aitis$^2$\\
$^1$Delft University of Technology, $^2$Microsoft\\

}
\end{center}
\setcounter{page}{1}

\section{Illustrating Tension Parameters}
\label{app:tension}
\begin{figure}[h]
\centering
\setkeys{Gin}{width=\linewidth}
\addtolength{\tabcolsep}{-0.6em}
\begin{tabularx}{\linewidth}{@{}*8{>{\centering\arraybackslash}X}@{}}
\includegraphics{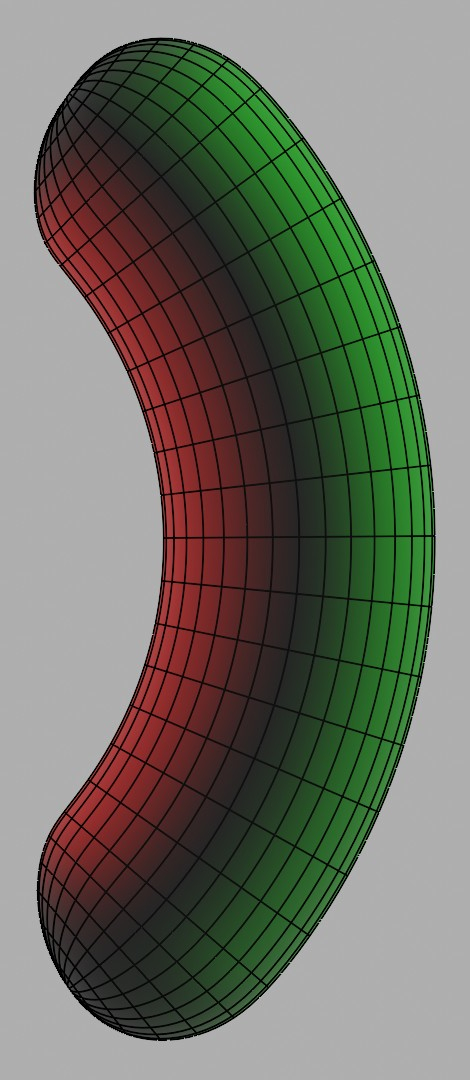} & 
\includegraphics{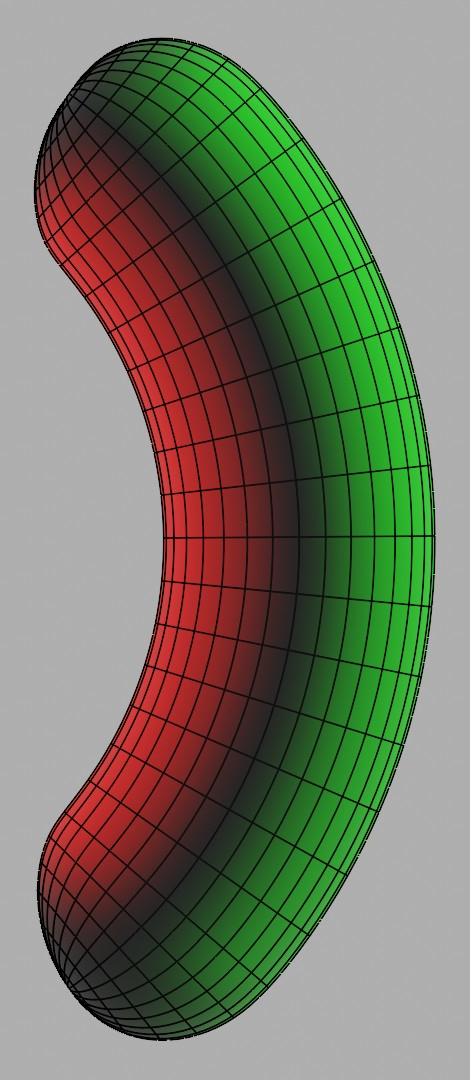} & 
\includegraphics{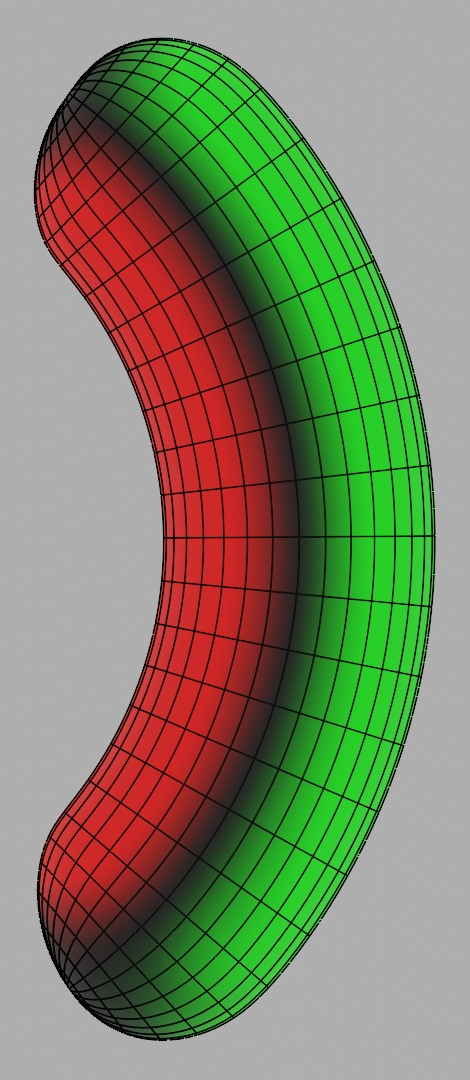} & 
\includegraphics{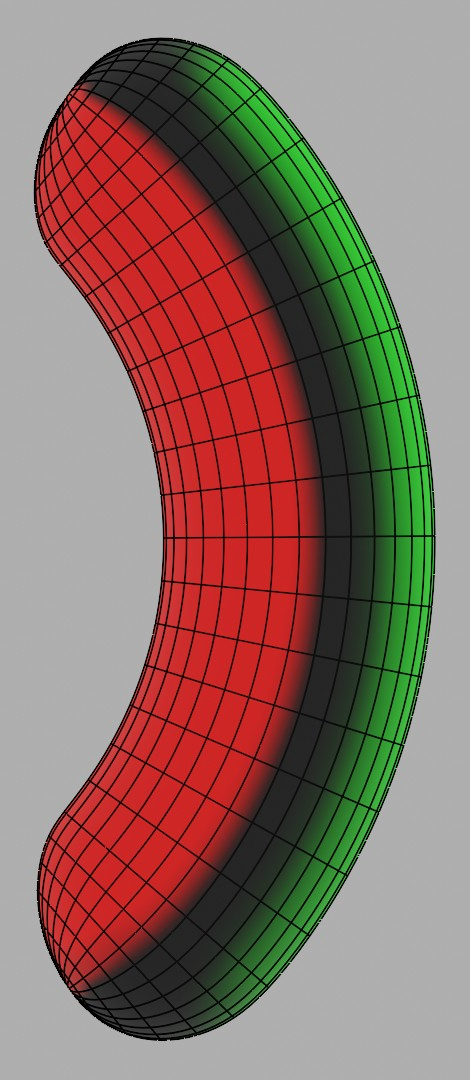} & 
\includegraphics{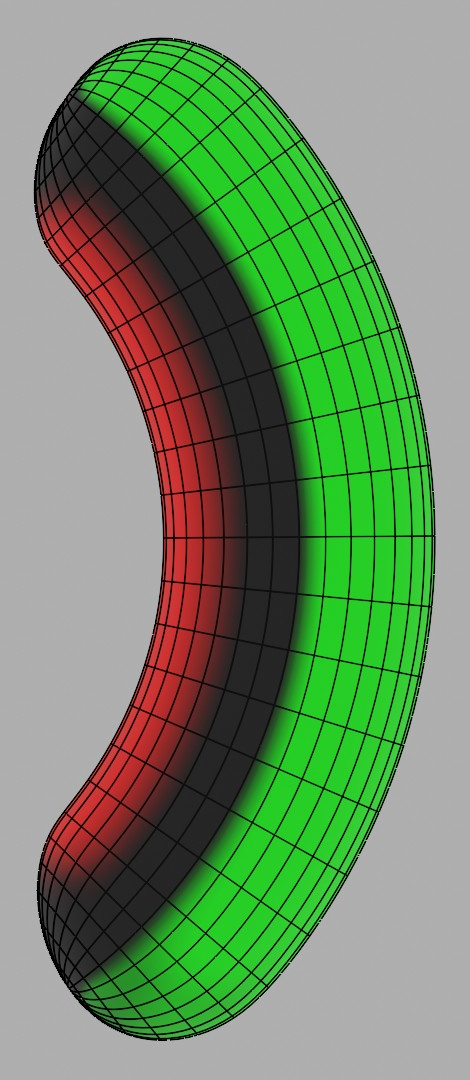} &
\includegraphics{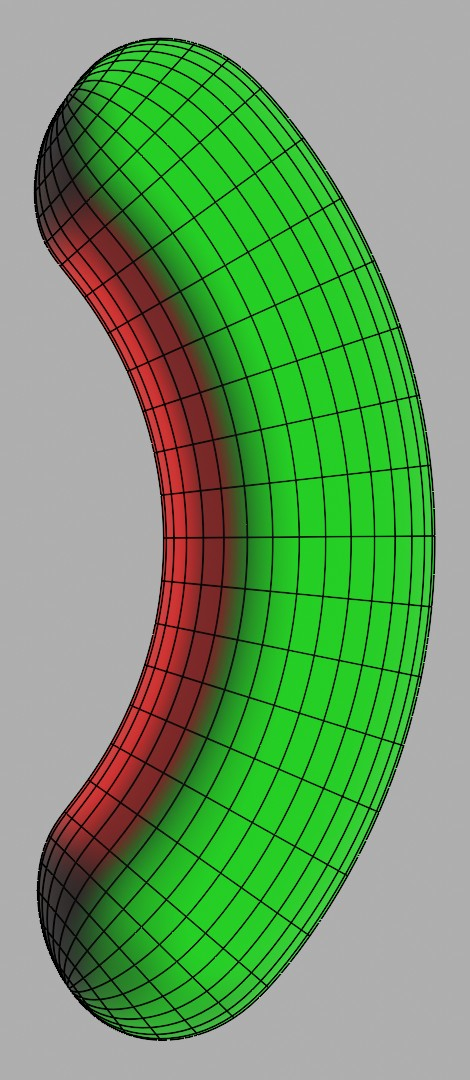} &
\includegraphics{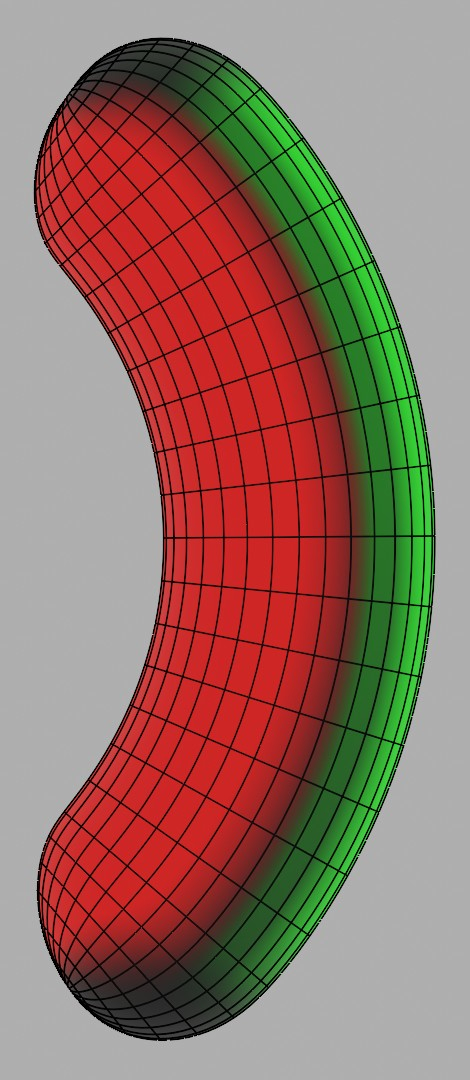} &
\includegraphics{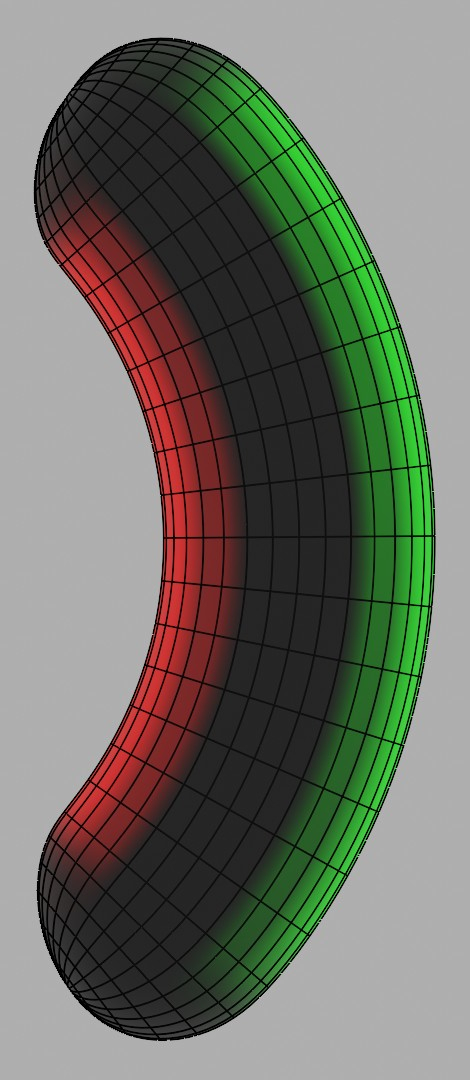}\\
\textsf{\scriptsize s: 3\quad b: 0} & \textsf{\scriptsize s: 5\quad b: 0} & \textsf{\scriptsize s: 10\quad b: 0} & \textsf{\scriptsize s: 10\quad b: -0.6} & \textsf{\scriptsize s: 10\quad b: 0.6} &
\textsf{\scriptsize s: 10\quad b: 0} &
\textsf{\scriptsize s: 10\quad b: 0} &
\textsf{\scriptsize  s: 10\quad b: 0}\\
\textsf{\scriptsize e: 0\quad c: 0} & \textsf{\scriptsize e: 0\quad c: 0} & \textsf{\scriptsize e: 0\quad c: 0} & \textsf{\scriptsize e: 0\quad c: 0} & \textsf{\scriptsize e: 0\quad c: 0} &
\textsf{\scriptsize e: 3\quad c: -3} &
\textsf{\scriptsize e: -3\quad c: 3} &
\textsf{\scriptsize  e: -3\quad c: -3}\\
\end{tabularx}
\caption{
\textbf{Tension Parameters - Cylinder.} Illustrating the effect of varying tension parameters on a simple cylinder mesh. Legend: \textit{s} - strength, \textit{b} - bias, \textit{e} - iterations for dilating/eroding expansion, \textit{c} - iterations for dilating/eroding compression.
}
\label{fig:tension-params-hotdog}
\end{figure}
\begin{figure*}[h]

\setkeys{Gin}{width=\linewidth}
\addtolength{\tabcolsep}{-0.5em}

\begin{minipage}{0.48\linewidth}
\begin{tabularx}{\linewidth}{@{}*3{>{\centering\arraybackslash}X}@{}}
\multicolumn{3}{c}{\textsf{Strength}} \\
\textsf{\footnotesize{Low}} & \textsf{\footnotesize{Default}} & \textsf{\footnotesize{High}} \\
\includegraphics{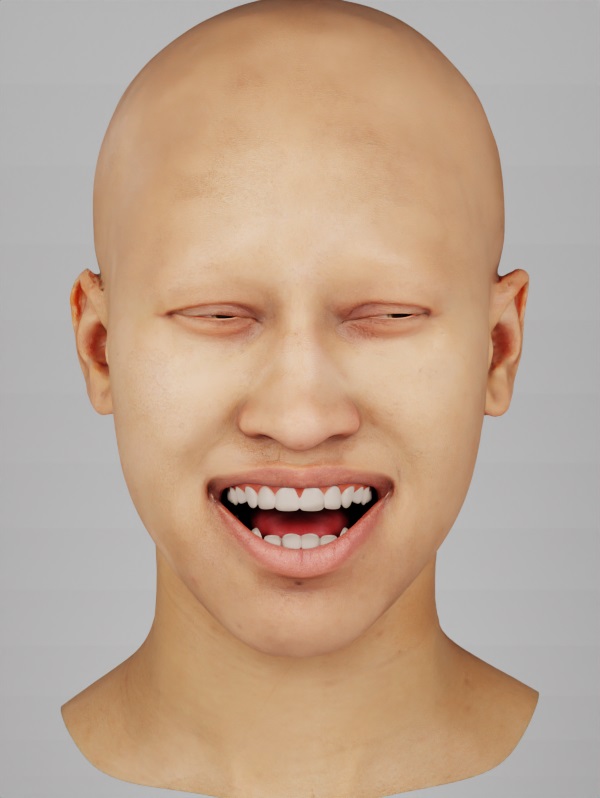} & 
\includegraphics{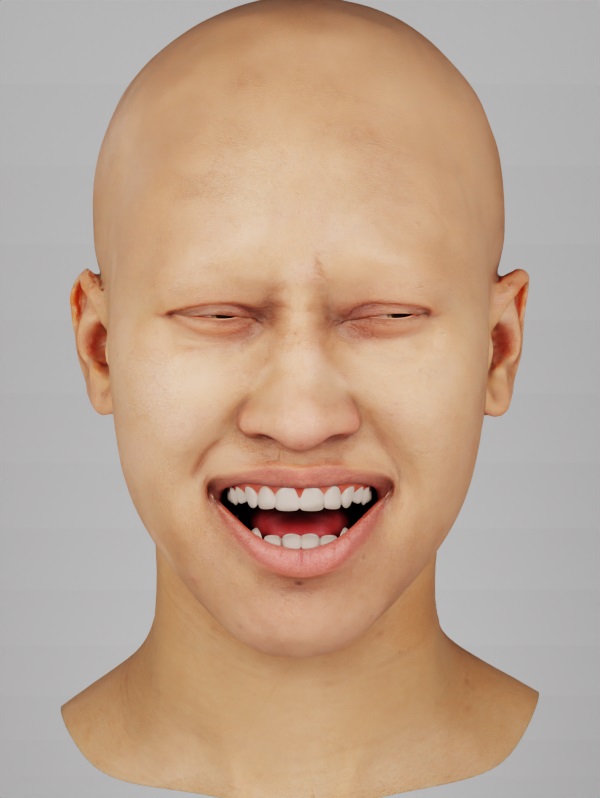} & 
\includegraphics{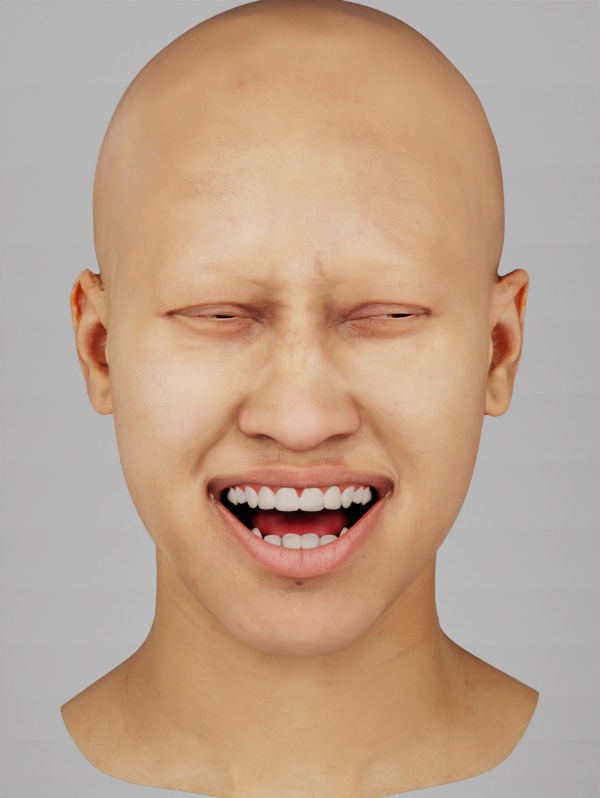} \\
\includegraphics{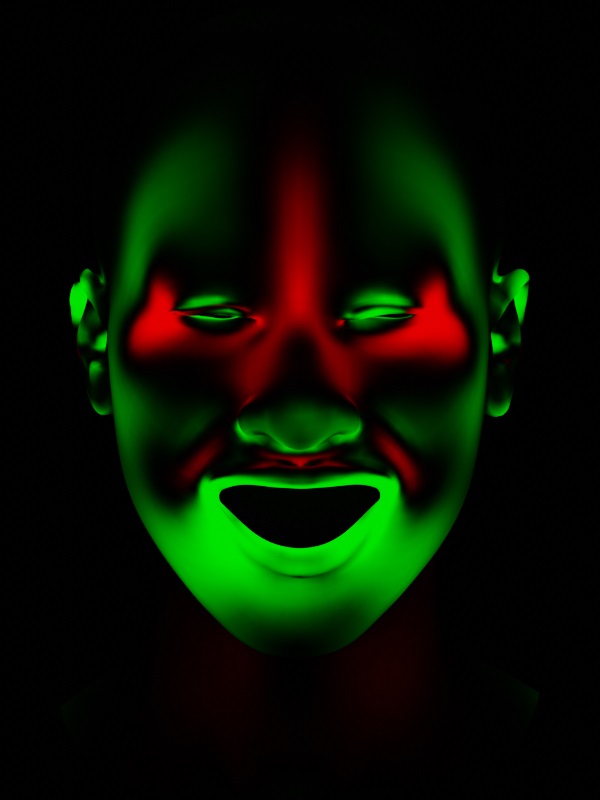} & 
\includegraphics{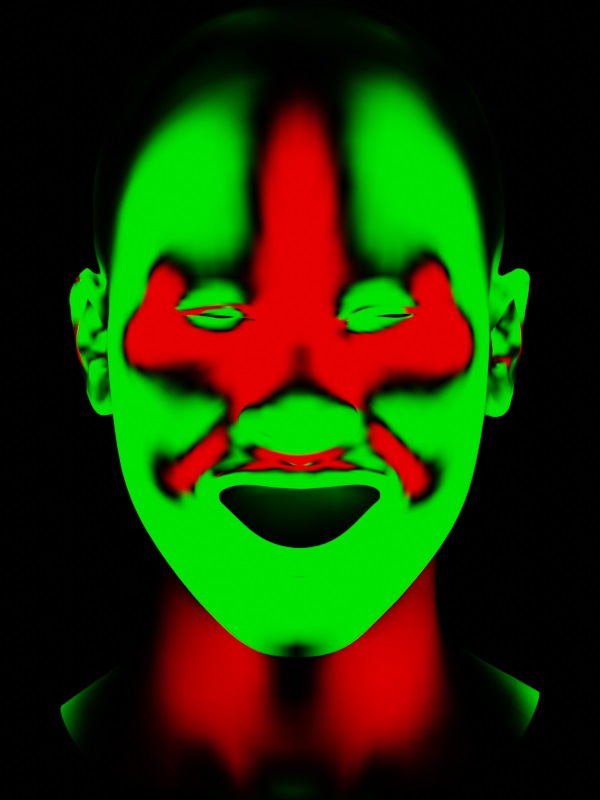} & 
\includegraphics{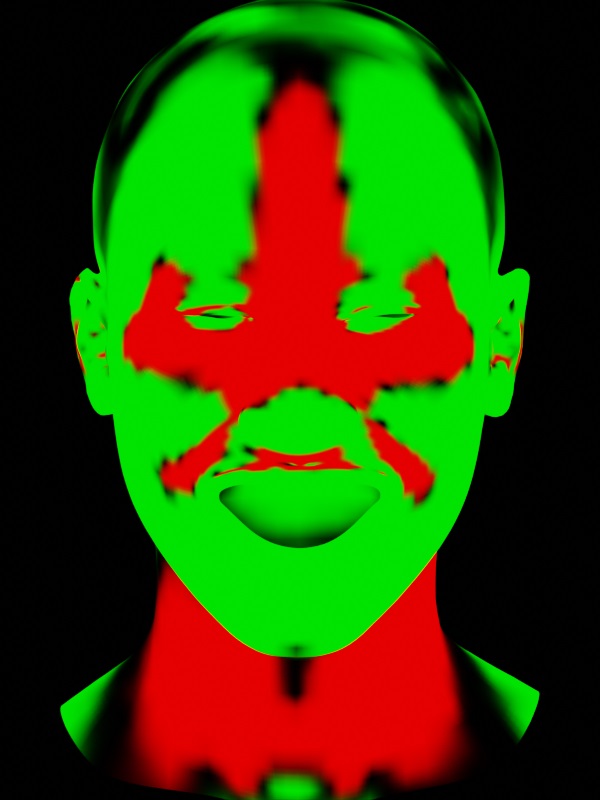}
\end{tabularx}
\end{minipage}
\hfill
\begin{minipage}{0.48\linewidth}
\begin{tabularx}{\linewidth}{@{}*3{>{\centering\arraybackslash}X}@{}}
\multicolumn{3}{c}{\textsf{Iterations}} \\
\textsf{\footnotesize{Low}} & \textsf{\footnotesize{Default}} & \textsf{\footnotesize{High}} \\
\includegraphics{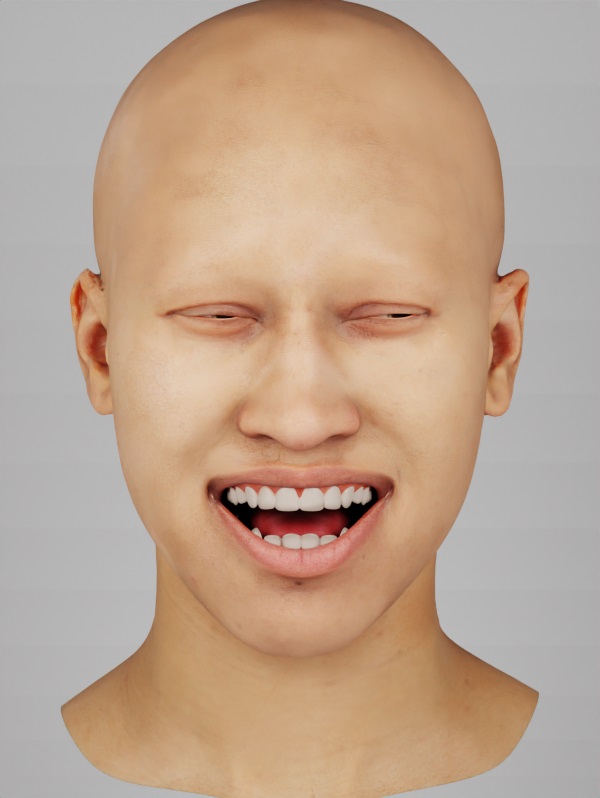} & 
\includegraphics{img/tension/vis/def.jpg} & 
\includegraphics{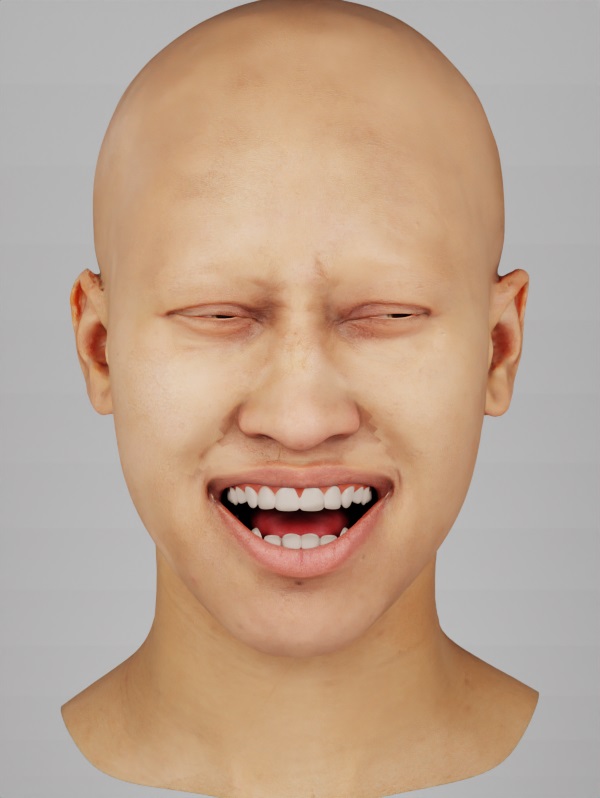} \\
\includegraphics{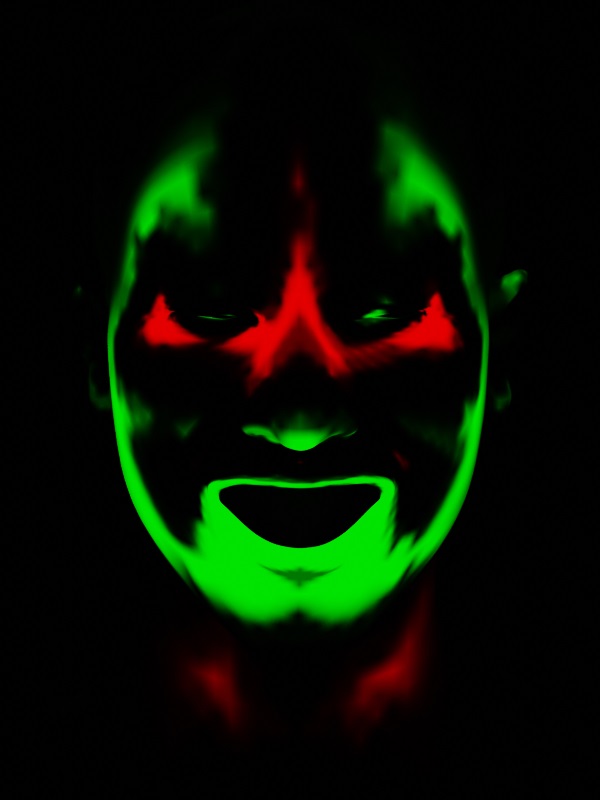} & 
\includegraphics{img/tension/ten/def.jpg} & 
\includegraphics{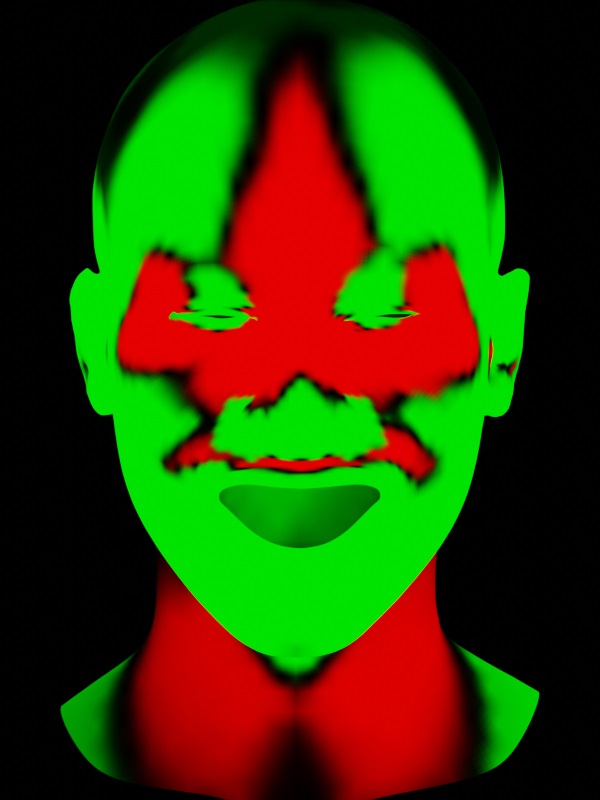}
\end{tabularx}
\end{minipage}

\caption{
\textbf{Tension Parameters - Face.} Illustrating the effect of varying tension parameters on a face mesh.
}
\label{fig:tension_params}
\end{figure*}
\clearpage

\section{Eye-Region Landmark Metrics}
\label{app:metrics}
To deal with different landmark annotation conventions (e.g. $68$, $98$, $703$ landmarks), we use a point to polyline distance. For each eyelid point in the prediction, we measure its distance to the relevant polyline, e.g. for a predicted point on upper-left eyelid we measure the distance from it to upper-left eyelid polyline (illustrated in \autoref{fig:point-to-line}).
This allows us to compare models with different annotation schemes and to have a better understanding of eye region error.

In cases where we do not have landmark annotations, but we know that both eyes are closed (or a single eye is closed), we can use the eye opening/aperture error instead.
This is illustrated in \autoref{fig:eye-open}.
The limitation of this approach is that it measure the relative openness of eye only and will have a low error even if the location of eyelid is wrong (but the aperture is correct).
However, in combination with other metrics it provides a good signal to how well the models deal in detecting winks and blinks.

\begin{figure}[!h]
  \begin{minipage}[t]{0.49\textwidth}
    \input{fig/point_to_line}
  \end{minipage}
  \hfill
  \begin{minipage}[t]{0.49\textwidth}
    \input{fig/eye_open}
  \end{minipage}
\end{figure}

\section{Landmark Predictions on 300W}
\label{app:300W-preds}

\begin{figure}
\renewcommand\tabularxcolumn[1]{m{#1}}
\centering
\begin{tabularx}{0.76\linewidth}{@{}*5{>{\centering\arraybackslash}X}@{}}
\textsf{\small Input Image} & 
\textsf{\small AWING \cite{wang2019adaptive}} & 
\textsf{\small 3FabRec \cite{browatzki2020}} & 
\textsf{\small No Wrinkles~\cite{woodFakeItTill2021}} & 
\textsf{\small Ours (Wrinkles)}\\
\end{tabularx}
\includegraphics[width=0.78\linewidth]{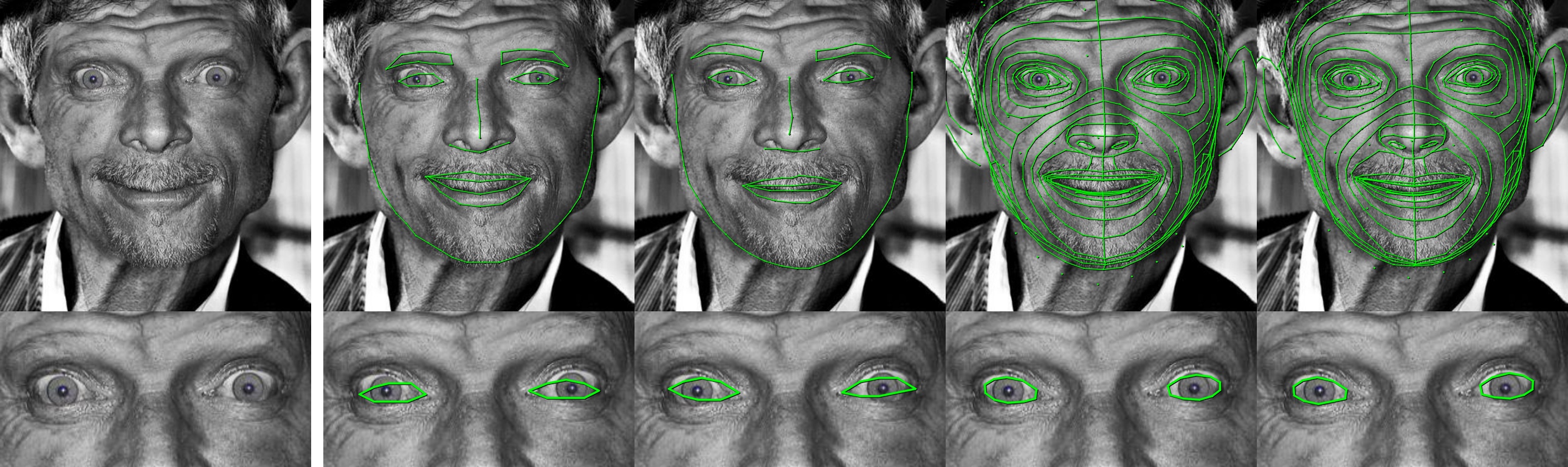}
\includegraphics[width=0.78\linewidth]{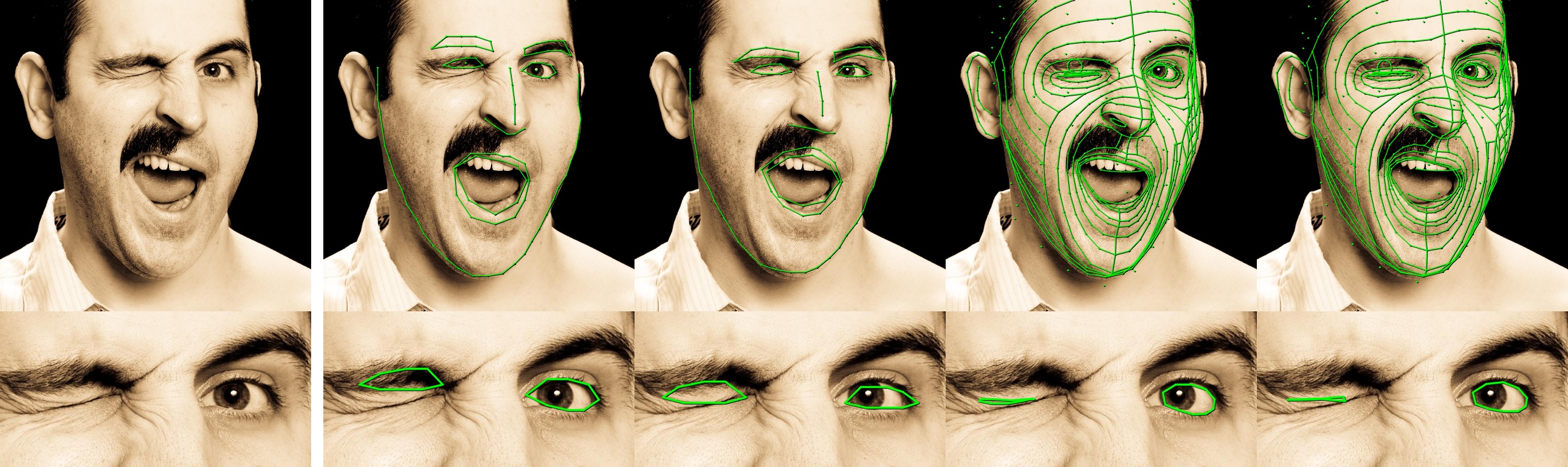}
\includegraphics[width=0.78\linewidth]{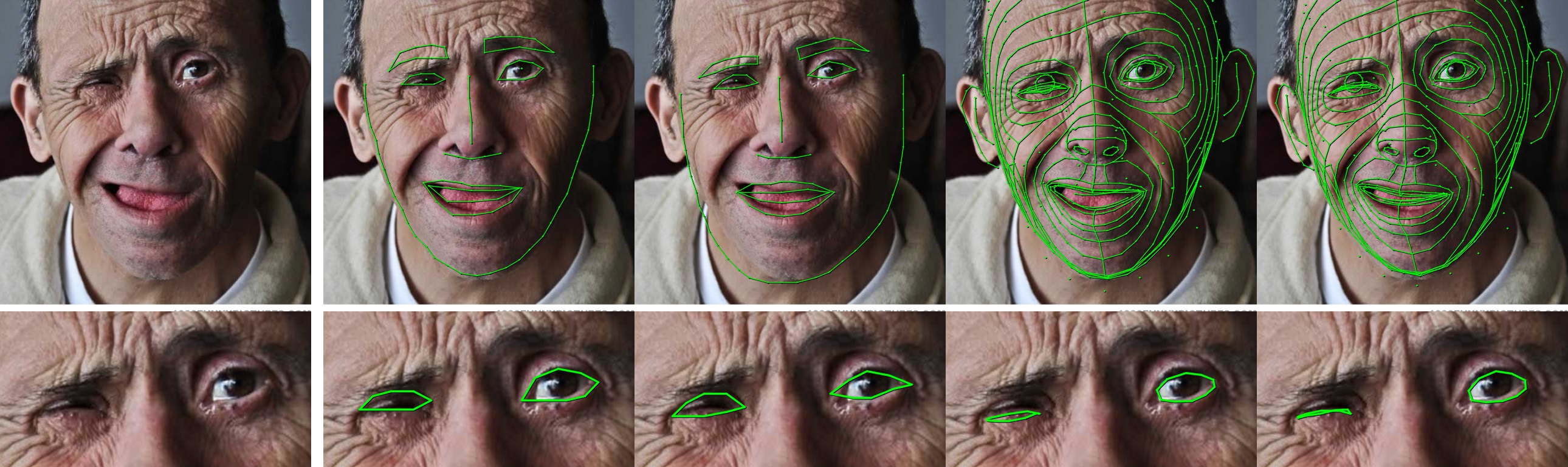}
\includegraphics[width=0.78\linewidth]{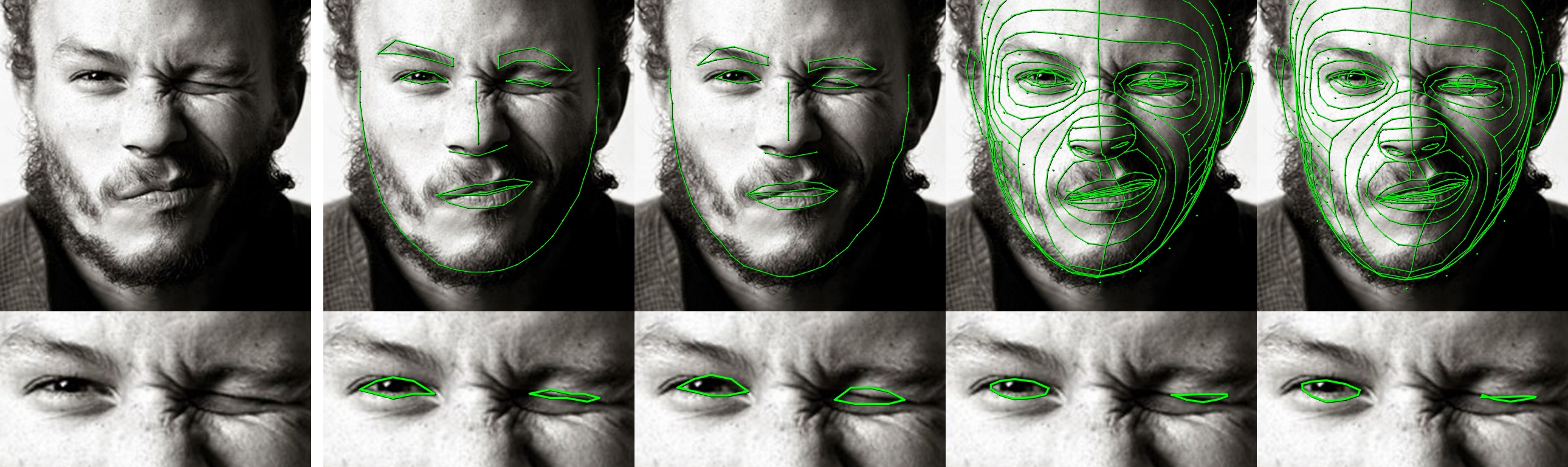}
\includegraphics[width=0.78\linewidth]{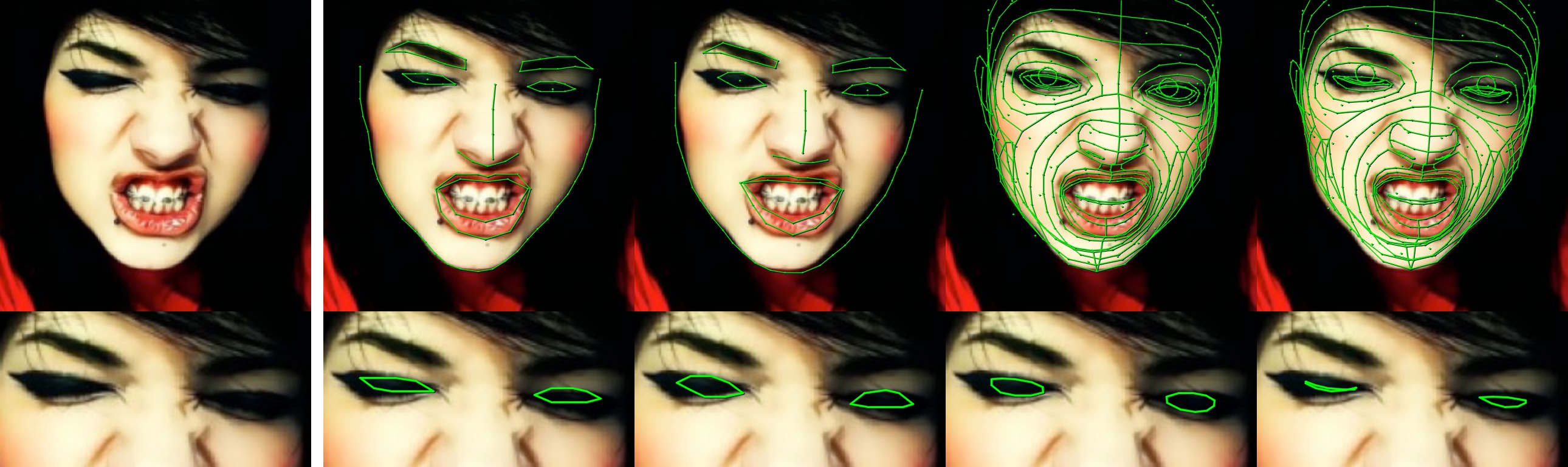}
\caption{
\textbf{Qualitative Comparison against SOTA}. Comparing prediction on 300W against SOTA models for facial landmark detection, our synthetic-only model often results in better accuracy for eye region, with improved performance for wink detection with mesh-based tension data.}
\label{fig:300W-sota}
\end{figure}

We present examples of predictions on 300W dataset from models trained on real and synthetic data in \autoref{fig:300W-sota}.

\section{Surface-Normals Predictions}
\label{app:normals}

\begin{figure*}
\renewcommand\tabularxcolumn[1]{m{#1}}
\centering
\begin{tabularx}{0.83\linewidth}{@{}*5{>{\centering\arraybackslash}X}@{}}
\textsf{\small Input Image} & 
\textsf{\small CMDFN~\cite{abrevaya2020cross}} & 
\textsf{\small SfSnet~\cite{sengupta2018sfsnet}} & 
\textsf{\small No Wrinkles~\cite{woodFakeItTill2021}} & 
\textsf{\small Ours (wrinkles)}
\end{tabularx}
\includegraphics[width=0.84\linewidth]{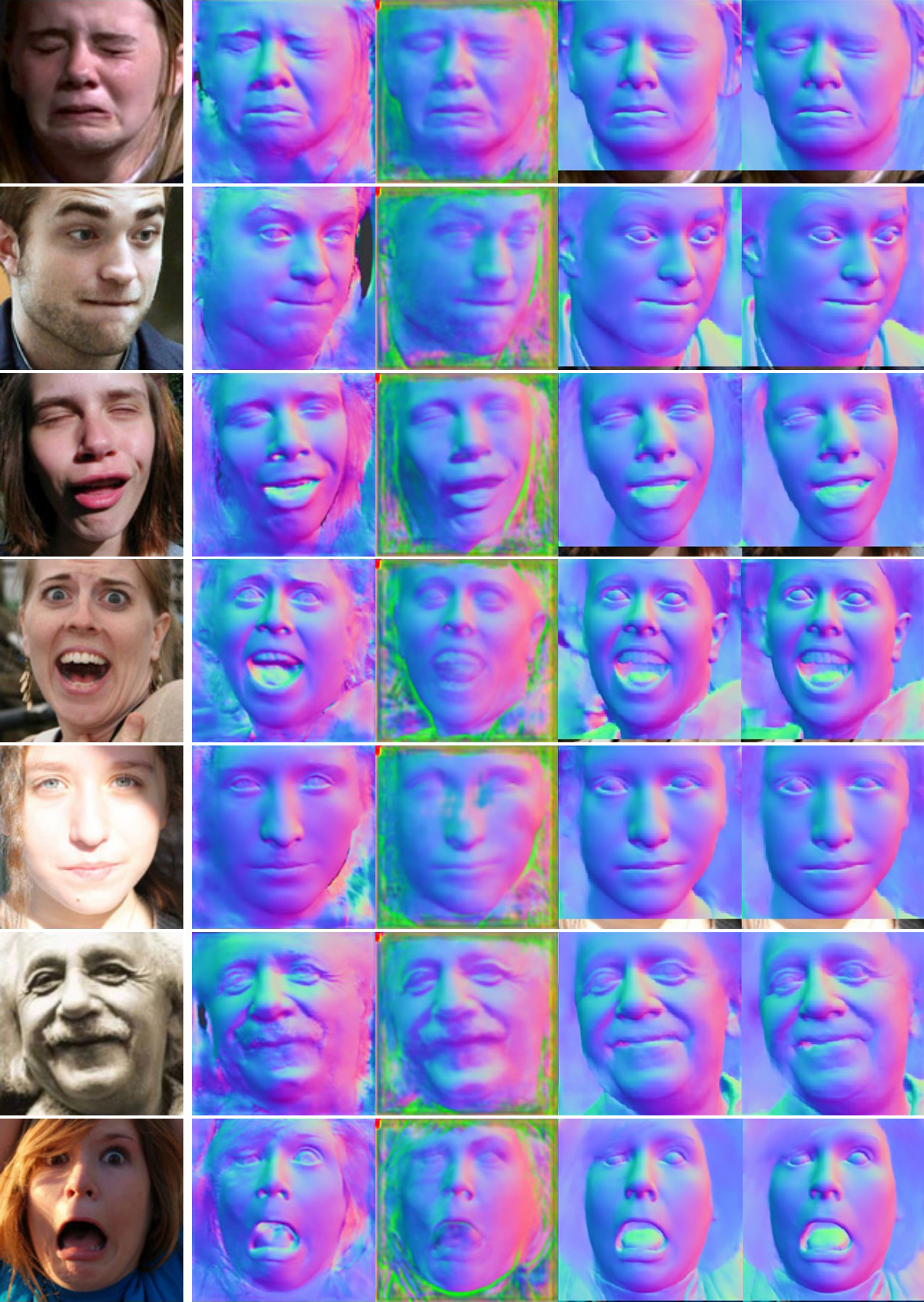}
\caption{
\textbf{Surface-Normals Comparison.} Qualitative results for surface-normals compared with two recent approaches. Note that we use our own face-alignment resulting in slight offset of the predicted ROI.
}
\label{fig:normals-supp}
\end{figure*}

In \autoref{fig:normals-supp} we show further comparisons to the recent face surface-normals prediction techniques of \citet{abrevaya2020cross} and \citet{sengupta2018sfsnet}.

\begin{figure}
\centering
\includegraphics[width=\linewidth]{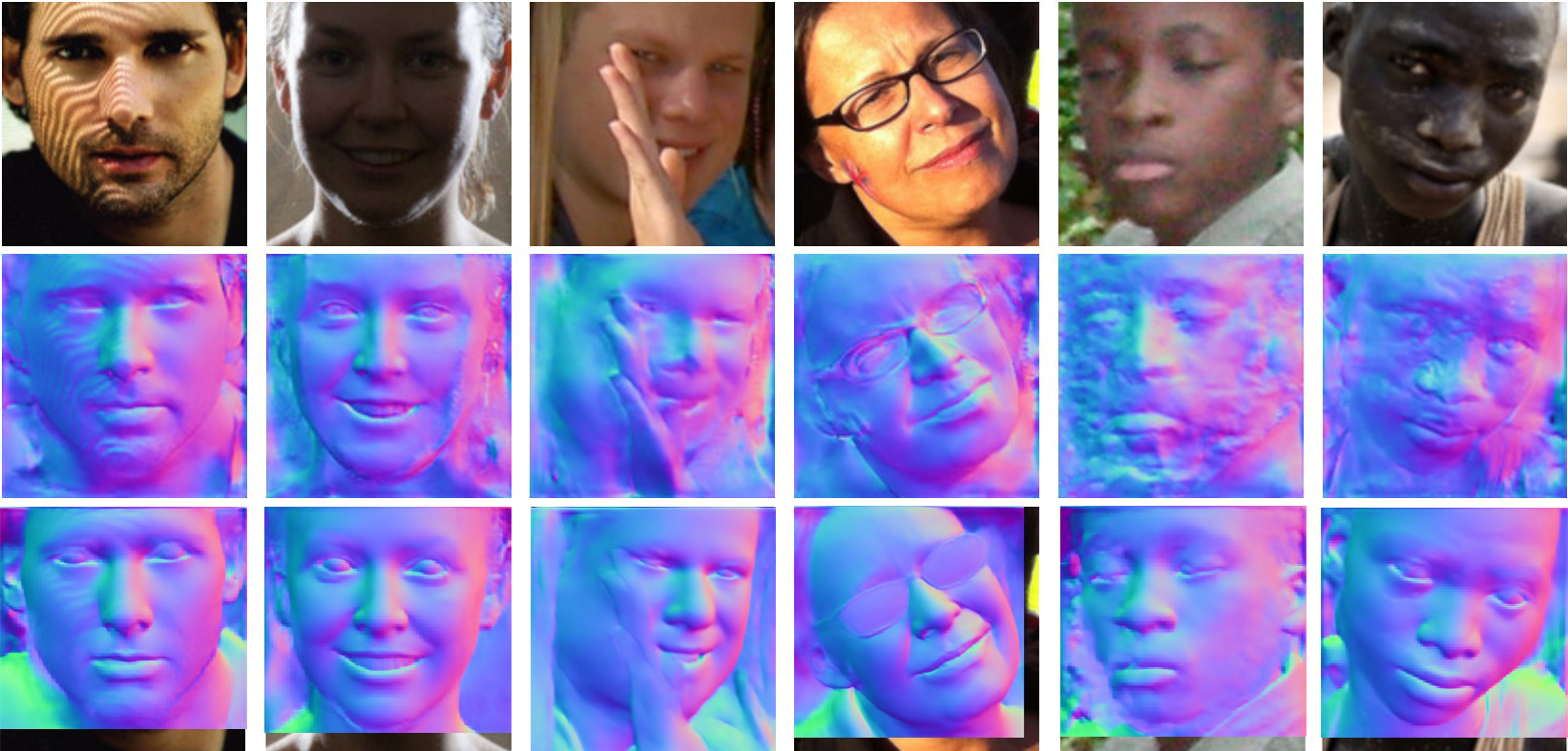}
\caption{
\textbf{Surface-Normals Robustness.} Comparison of surface-normals prediction to failure cases from Abrevaya et al.~\cite{abrevaya2020cross}, showing input (top), Abrevaya et al. (middle) and Ours (bottom). Our approach is significantly more robust in cases of extreme lighting, occlusion and darker skin.
}
\label{fig:normals-comp}
\end{figure}

\autoref{fig:normals-comp} shows failure cases from \citet{abrevaya2020cross} and our results on the same images. 
It is clear that our technique results in a significantly more robust model which can deal better with extreme lighting conditions, occlusions and darker skin tones.
Note that when training our model we take the surface of glasses lenses into account, though it is also possible to ignore these and predict for the face underneath depending on how the rendering pipeline is configured.
In all figures relating to surface-normals prediction we use our own face alignment to select the region of interest (ROI) to input to the normals prediction U-Net model, which causes some misalignment with the ROI used in other techniques.

\section{The 300W-winks Subset}
\label{app:300w-winks}

The subset of 300W images that make up 300W-winks is:
\\
\texttt{\small indoor\_048, indoor\_052, indoor\_053, indoor\_054, indoor\_055, indoor\_089, indoor\_094, indoor\_099, indoor\_180, indoor\_226, indoor\_242, indoor\_253, indoor\_264, indoor\_267, indoor\_278, indoor\_280, indoor\_282, indoor\_286, outdoor\_073, outdoor\_076, outdoor\_077, outdoor\_089, outdoor\_097, outdoor\_145, outdoor\_165, outdoor\_209, outdoor\_243, outdoor\_249, outdoor\_251, outdoor\_292}.

\section{The Pexels Winks and Blinks Dataset}
\label{app:pexels}

All images can be accessed by appending to \url{https://www.pexels.com/photo/}

\subsection{Blinks subset}
\noindent\texttt{\small abhishek-sinari-2026945, adrienne-andersen-2552127, adrienne-andersen-2552131, alekke-blazhin-7448048, alekke-blazhin-8450287, alekke-blazhin-8450288, alekke-blazhin-8450290, alekke-blazhin-8450296, alena-darmel-6940463, alena-shekhovtcova-7036537, alesia-kozik-7295537, alex-green-6626087, alexander-krivitskiy-4383786, alexander-stemplewski-2906663, alexandr-podvalny-1540408, alexandra-patrusheva-6806789, ali-karimiboroujeni-11381826, alina-blumberg-6925493, alyona-pastukhova-11495069, amar-preciado-10820282, amr-osman-10665375, anastasia-ilinamakarova-10832112, anastasia-shuraeva-7539962, anastasia-trofimczyk-10311002, anastasiia-chaikovska-11834502, anastasiia-shevchenko-10568846, andre-porto-7753232, andrea-gulotta-11140482, andrea-piacquadio-3757942, andrea-piacquadio-3760262, andrea-piacquadio-3760611, andrea-piacquadio-3764535, andrea-piacquadio-3768163, andrea-piacquadio-3768724, andrea-piacquadio-3771813, andrea-piacquadio-3786522, andrea-piacquadio-3799096, andrea-piacquadio-3799787, andrea-piacquadio-3799830, andrea-piacquadio-3807762, andrea-piacquadio-3811603, andrea-piacquadio-3811663, andrea-piacquadio-3812746, andrea-piacquadio-3831645, andrea-piacquadio-941693, andres-ayrton-6578880, anete-lusina-4793357, anete-lusina-5723189, angela-roma-7479819, angelica-reyn-11893387, anh-tuan-9889769, anna-shvets-3746281, anna-shvets-3852192, anna-shvets-4557467, anna-shvets-4611655, anna-shvets-4971107, anna-shvets-5034475, anna-shvets-5069470, anna-shvets-5069493, anna-shvets-5069609, anna-tarazevich-5155727, anna-zaykina-8452431, antoni-shkraba-5890702, antoni-shkraba-7484863, arianna-jade-2896823, arina-krasnikova-6663361, arina-krasnikova-6663367, arina-krasnikova-6914826, arina-krasnikova-6914833, arina-krasnikova-6998572, arina-krasnikova-7752573, arina-krasnikova-7752693, armin-rimoldi-5269495, arsham-haghani-3423024, artyom-malyukov-11896104, azraq-al-rezoan-11763863, azraq-al-rezoan-11763868, barathan-amuthan-2723624, ben-mack-6775289, blue-bird-7210441, breno-santos-10060165, brett-sayles-4095246, caique-araujo-10218049, camilla-gari-10306657, charles-wundengba-3609781, cliff-booth-4057336, cottonbro-10049355, cottonbro-10140838, cottonbro-10678800, cottonbro-4727484, cottonbro-5020308, cottonbro-5386370, cottonbro-5561559, cottonbro-5561563, cottonbro-5850831, cottonbro-5976145, cottonbro-6700116, cottonbro-6700119, cottonbro-6700142, cottonbro-6700144, cottonbro-6753360, cottonbro-6753370, cottonbro-6753371, cottonbro-7407129, cottonbro-8102360, cottonbro-8142260, cottonbro-9063608, cottonbro-9063624, cottonbro-9063626, cottonbro-9316296, cottonbro-9467199, cottonbro-9577189, cottonbro-9955927, craig-adderley-2306203, craig-adderley-2306210, craig-adderley-2306213, cup-of-couple-6634443, cup-of-couple-6962575, cup-of-couple-6963527, daria-nekipelova-9665517, daria-rem-1977055, darina-belonogova-7886748, darina-belonogova-8386475, davner-toledo-4574403, dziana-hasanbekava-6851631, efigie-lima-marcos-11831324, ehsan-7538807, ekaterina-bolovtsova-7113346, ekaterina-bolovtsova-7113362, elina-fairytale-3865731, elina-fairytale-3865763, elina-fairytale-3865765, eman-genatilan-5348809, eman-genatilan-8589781, emmy-pua-10196907, engin-akyurt-5059305, eric-deine-11781294, estelle-umaes-11734787, evelina-zhu-6286063, faruk-tokluoglu-8777603, fireberrytech-6683091, flint-huynh-11804619, gary-barnes-6248993, gary-barnes-6249024, greta-hoffman-7675722, guilherme-almeida-1858175, hebert-santos-5485599, ichad-windhiagiri-7616249, imad-clicks-11742222, imad-clicks-11742223, ivan-mudruk-10400317, ivan-samkov-6968814, ivan-samkov-8952728, jamie-saw-10029674, jeandaniel-francoeur-7678688, jennifer-enujiugha-1904674, jill-burrow-6758033, joao-vitor-heinrichs-1787039, joshua-abner-3605015, joshua-mcknight-3290242, julia-avamotive-1070967, julia-tatyanenko-11855943, juliana-marinina-9957288, kampus-production-6298293, kampus-production-6298321, kampus-production-7928134, kampus-production-8871412, karolina-grabowska-4378486, karolina-grabowska-4498195, kat-smith-568021, ketut-subiyanto-4473864, ketut-subiyanto-4545165, ketut-subiyanto-4584390, ketut-subiyanto-4584601, kindel-media-7298396, kindel-media-7298459, kindel-media-7938549, kirill-palii-3545783, klaus-nielsen-6303717, korede-adenola-11785507, kseniya-buraya-10008858, kwesiblaq-10986569, leah-kelley-3722162, leo-acevedo-3261142, lucas-souza-1964442, lucas-souza-3608010, maksim-goncharenok-4892914, marcelo-chagas-2535859, maria-eduarda-loura-magalhaes-4340053, maria-luiza-melo-11819746, maria-orlova-4946635, maria-orlova-4947740, marija-7737766, marlon-schmeiski-11193234, mart-production-7880131, matheus-bertelli-11749497, matheus-ferrero-11470717, matheus-henrin-11360455, maycon-marmo-4346013, michelle-leman-6774345, mike-cabugao-8503888, mikhail-nilov-6707031, mikhail-nilov-6943956, mikhail-nilov-6945088, mikhail-nilov-6968191, mikhail-nilov-6968331, mikhail-nilov-7776528, mikhail-nilov-8343016, mikhail-nilov-8350479, ming-zimik-5861623, miriam-alonso-7623727, monica-turlui-8218377, monstera-5063295, monstera-5273734, monstera-5302897, monstera-5384518, monstera-6781240, monstera-6973715, monstera-6974031, monstera-6977869, monstera-7352909, mosei-films-9209576, nadin-sh-11872307, nguyen-phuong-linh-6211165, nicola-barts-7925781, nikita-nikitin-11008044, nikita-semezhin-9787604, oleg-magni-1669154, olia-danilevich-8964938, olya-prutskova-7179057, orione-conceicao-2983464, ozan-culha-11850759, ozan-culha-11858978, ozan-culha-11866492, pavel-danilyuk-7267691, pavel-danilyuk-7267700, pavel-danilyuk-7406040, pexels-user-9281097, pnw-production-8980983, pnw-production-8981313, polina-chistyakova-9052464, polina-kovaleva-5885655, polina-kovaleva-7090394, polina-tankilevitch-6630835, polina-tankilevitch-8210939, rachel-claire-4992586, rafael-freire-5714746, rafael-portraits-9281360, raquel-silva-11870922, renthel-cueto-11131698, renthel-cueto-11131703, rfstudio-3843292, rheyan-glenn-dela-cruz-manggob-10210334, rodnae-productions-7402945, rodnae-productions-8173525, rodnae-productions-8173543, roman-odintsov-11760366, roman-odintsov-11760376, roman-odintsov-11760378, roman-odintsov-8018975, ron-lach-10139616, ron-lach-10321431, ron-lach-8159655, run-ffwpu-11757051, ruslan-rozanov-11585357, samer-daboul-4506967, samson-katt-5256085, santiago-jose-calvo-11757764, sasha-lazarev-3578326, shiny-diamond-3762659, shotpot-6338298, shvets-production-6974955, shvets-production-6975262, shvets-production-6975383, shvets-production-6975413, shvets-production-6984635, shvets-production-8005151, si-luan-pham-8778439, sound-on-3756943, svetlana-10311383, taina-bernard-3482526, tanya-gorelova-3855199, thiago-alencar-10154765, thiago-matos-10359136, thirdman-6958390, thirdman-7237074, thirdman-7268229, thirdman-7268234, thirdman-7268483, thirdman-8053704, thomas-nguka-10163670, thomas-nguka-7562643, tieu-bao-truong-8298108, tiffany-freeman-11038435, tima-miroshnichenko-5118496, tima-miroshnichenko-6670752, tubarones-photography-2737046, tubarones-photography-2943689, tubarones-photography-3065450, valdemar-9546870, vanessa-loring-5082946, vika-kirillova-10119334, vika-kirillova-11067905, vinicius-altava-2657594, vitoria-santos-1913161, vitoria-santos-2838831, vlada-karpovich-8528898, vlada-karpovich-8939842, vladimir-konoplev-11323367, vladimir-konoplev-11323376, vladimir-vasilev-7640302, wesley-carvalho-4126255, yan-krukov-6617027, yan-krukov-7155545, yana-sperry-11810044, yaroslav-shuraev-6281021, zayceva-tatiana-11210581, zayceva-tatiana-11698072}

\subsection{Winks subset}
\noindent\texttt{\small airam-datoon-9637814, alena-darmel-7322312, alena-darmel-8153597, alexander-krivitskiy-6471731, alexander-krivitskiy-6828450, amina-filkins-5560027, amina-filkins-5560029, amina-filkins-5561443, amina-filkins-5561455, andrea-piacquadio-3764391, andrea-piacquadio-3777558, andrea-piacquadio-3777563, andrea-piacquadio-3778216, andrea-piacquadio-3778673, andrea-piacquadio-3779420, andrea-piacquadio-3783107, andrea-piacquadio-3979192, arina-krasnikova-6663385, beatriz-braga-10461875, bia-sousa-2191056, brianna-amick-2069008, bruno-ticianelli-1889787, chandrashekar-hosakere-matt-707449, cleyder-duque-3690938, cottonbro-5416365, cottonbro-6144420, darija-shelkovich-5010665, dazzle-jam-2020992, debendra-das-5429301, deepak-digwal-3577006, ekaterina-glushenko-8993965, evgeny-zuchman-9986411, furkanfdemir-7020329, ganeshbabu-arun-580012, gustavo-fring-4254148, gustavo-fring-6050330, gustavo-fring-7447022, ike-louie-natividad-3208616, jekaterina-glushenko-8993965, jevgenija-shuhman-9986411, john-valette-10785650, julia-larson-6113247, julia-larson-6113626, julia-larson-6113631, julia-larson-6113639, julia-larson-6113641, kaushal-moradiya-3400573, kenneth-gorzal-surillo-8177834, ketut-subiyanto-4584180, ketut-subiyanto-4584197, ketut-subiyanto-4584387, ketut-subiyanto-4584604, koolshooters-6976854, koolshooters-7142955, koolshooters-7143075, koolshooters-7143155, koolshooters-7143217, ksenia-chernaya-6616201, laura-tancredi-7065531, matheus-bertelli-7510832, melissa-jansen-van-rensburg-2255462, mikhail-nilov-8922330, mikhail-nilov-8922333, mikhail-nilov-8923584, monstera-7114452, monstera-7114634, monstera-7139819, nataliya-vaitkevich-4813824, nataliya-vaitkevich-4813859, nichole-sebastian-3264351, nikita-saif-11494236, pavel-danilyuk-8422489, pavel-danilyuk-8422507, pavel-danilyuk-8422515, pixabay-40565, polina-tankilevitch-4723521, polina-tankilevitch-4723538, polina-tankilevitch-4725082, polina-tankilevitch-4725084, polina-tankilevitch-4725108, polina-tankilevitch-4725153, polina-tankilevitch-6630839, polina-tankilevitch-6988592, polina-zimmerman-3958873, rahmi-aksoz-9957220, rich-ortiz-5661730, rodnae-productions-10503462, rodnae-productions-6709127, ron-lach-8159657, ron-lach-8989996, sarah-chai-7262397, shvets-production-6975619, shvets-production-7525145, sora-shimazaki-5938614, th-team-7516292, the-weddingfog-9084064, thirdman-6109560, victoria-strelkaph-11034423, yan-krukov-4964933, yan-krukov-4964936, yan-krukov-4964937, yan-krukov-7793112, yuliya-shabliy-388517, zura-modebadze-4922053, zura-modebadze-4922064}

\end{document}